\pdfoutput=1
\documentclass{article} % For LaTeX2e
\usepackage{preprint,times}
\usepackage{xspace}
\newcommand{\method}{\mbox{PhySimCode}\xspace}
\newcommand{\model}{\mbox{PhySimCode}\xspace}

\usepackage{pifont}
\newcommand{\cmark}{\ding{51}}
\newcommand{\xmark}{\ding{55}}
\usepackage[table,dvipsnames]{xcolor}
\usepackage{subcaption}

\usepackage{amsmath,amsfonts,bm}

\def\eqref#1{equation~\ref{#1}}
\def\1{\bm{1}}

\DeclareMathAlphabet{\mathsfit}{\encodingdefault}{\sfdefault}{m}{sl}
\SetMathAlphabet{\mathsfit}{bold}{\encodingdefault}{\sfdefault}{bx}{n}

\usepackage{hyperref}
\usepackage{url}
\usepackage{graphicx}
\usepackage{algorithm} 
\usepackage{algorithmic}
\usepackage{booktabs}
\usepackage{listings}
\usepackage{adjustbox}
\usepackage{longtable}

\definecolor{promptbg}{RGB}{248,248,248}
\definecolor{promptframe}{RGB}{200,200,200}
\definecolor{promptkw}{RGB}{60,60,180}

\lstdefinestyle{prompt}{
  basicstyle      = \ttfamily\footnotesize,
  backgroundcolor = \color{promptbg},
  frame           = single,
  framesep        = 4pt,
  rulecolor       = \color{promptframe},
  breaklines      = true,
  breakatwhitespace = false,
  postbreak       = \mbox{\textcolor{red}{$\hookrightarrow$}\space},
  showstringspaces = false,
  numbers         = none,
  columns         = fullflexible,
  keepspaces      = true,
  upquote         = true,
  literate        = {>=}{{$\geq$}}1 {<=}{{$\leq$}}1 {!=}{{$\neq$}}1
}

\title{\model: A Benchmark and Evaluation Method for Physics Video to Code Generation}

\author{%
Souraja Kundu\textsuperscript{1}\thanks{Corresponding author (sourajak@cs.cmu.edu).} \quad
Aditya Gupta\textsuperscript{1} \quad
Joel Julin\textsuperscript{1} \quad
Yizhou Zhao\textsuperscript{1} \quad
Liuyue Xie\textsuperscript{2} \quad
Laszlo A. Jeni\textsuperscript{1} \\
\textsuperscript{1}Carnegie Mellon University
\\
\textsuperscript{2}Amazon AGI
}

\begin{document}

\maketitle

\begin{center}
Project page: \url{https://sourajakundu.github.io/PhySimCode/}
\end{center}

\begin{abstract}
Multimodal large language models (MLLMs) are good at video understanding and code generation, but it remains unclear whether they can recover the physical laws underlying videos, a capability essential for converting videos into executable simulations and for providing structured physical supervision in robotics. Captioning motion video does not imply that a model has inferred its governing dynamics, estimated quantitative parameters, or can regenerate the observed scene from code. Existing video–physics benchmarks evaluate plausibility, property estimation, or future-state prediction, but offer limited insight into whether models can translate visual dynamics into runnable simulation code. We introduce \textbf{\model}, a benchmark for physics video-to-code generation: given only a raw video, a model must infer the relevant physics, recover parameter values, and generate runnable Python that reproduces the input video. The dataset contains \textbf{160.6K procedurally generated samples} spanning \textbf{162 physics phenomena} across analytic SciPy simulations and PyBullet-rendered rigid-body, articulated, fluid, and contact-rich scenes; each sample bundles the source simulation code, sampled physical parameters, rendered video, and a structured chain-of-thought covering observations, laws, ODEs, numerical methods, and validation checks. We evaluate ten state-of-the-art closed- and open-source MLLMs along physical-law correctness, simulation equivalence, code executability, and parameter recovery. Models reliably state physics laws (4.27–4.93/5) yet score only 1.51–2.47/5 on inferring them from video, and the best system (Claude Sonnet 4.6) recovers just 7.7\% of ground-truth parameters within $\pm 20\%$, exposing a substantial video-to-simulation gap and positioning \model as a testbed for multimodal systems that translate perception into precise, executable, and scientifically verifiable physics. \href{https://huggingface.co/datasets/SourajaKundu123/PhySimCode}{The 2,430-video evaluation set} and \href{https://github.com/SourajaKundu/PhySimCode}{evaluation code} are released.
\end{abstract}
\section{Introduction}
\label{intro}

State-of-the-art multimodal large language models (MLLMs) increasingly perform well in image/video understanding and coding tasks, but do they fully understand what they see (like humans)? Think of video captioning; these models' internal representations of visual inputs limit precision and verifiability: raw pixel embeddings are dense but opaque, and natural-language captions are readable but lossy. For \textit{structured, information-dense, scientific} visuals: infographics and physics-simulation videos, captions fail to recover the dense numerical values or the spatiotemporal dynamics that a domain expert would immediately read off the input. Code, by contrast, is simultaneously human-readable, machine-executable, structured, and lossless with respect to such scientific visual input.

Recent work targets scientific chart-to-code generation~\citep{tang2026mm,zhao2025vincicoder,zhao2025chartcoder,wu2025plot2code}. An infographic's plotting code has two parts: (i) the \textbf{template}: chart type, colors, labels, legends, and (ii) the \textbf{data values} displayed. In a pie chart of monthly tourist counts, the template is the chart design (slices, colors, labels) and the values are the per-month counts; many code variants can render the same chart, but the template specs and the values are unique to that infographic.

A structurally analogous problem exists in the video domain: physics-simulation video-to-code generation. A physics-simulation video is the video counterpart of a structured, information-dense scientific image, and the same template/value decomposition applies: different libraries, physics engines, or procedural languages (C\#, Python) can produce identical dynamics, but the underlying physical laws and the values of the physical parameters (mass, length, elasticity, etc.) are unique. For example, a spinning disk shedding cubes from its rim involves rigid-body rotation, centripetal acceleration, the static-to-kinetic friction transition, and projectile motion under gravity. Because the same phenomenon can be expressed in different formulations (e.g.\ Newtonian or Lagrangian mechanics), we define the template at the level of the universal underlying physical concepts, such as conservation of energy or momentum.

\begin{figure}[t]
\centering
\includegraphics[width=\linewidth]{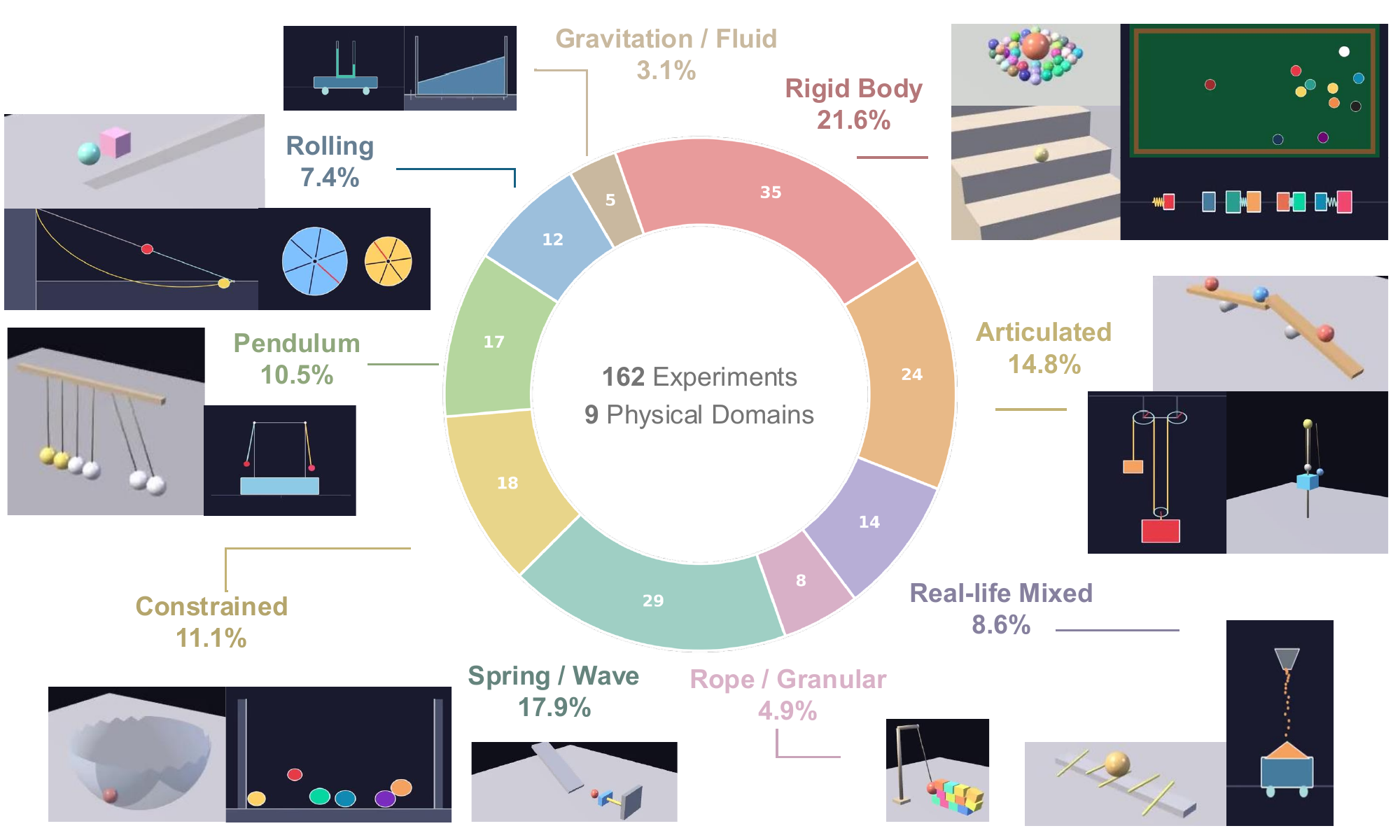}
\caption{Categorical distribution of physical domains with some Scipy (2D) and Kubric (3D) examples from our dataset.}
\vspace{-2mm}
\label{fig:pie}
\end{figure}

Once the template and parameter values are defined, we can test whether SOTA MLLMs identify both directly from the video: a direct probe of their video understanding and information-extraction capability. A stronger probe of an MLLM's \emph{true} physics understanding is to use those extracted concepts and values in a downstream task: generate simulation code that regenerates the exact same video. Because humans are naturally very sensitive to physical plausibility, executing the code and visually comparing the regenerated simulation to the original provides an intuitive and reliable evaluation: if the code is correct it serves as a lossless executable representation of the input video. Such physics-aware symbolic simulations of 2D/3D scenes are also valuable for teaching robots motion planning and the physics of the real world.

The task measures MLLM capability at three stages: (i) \textbf{video understanding}: extracting all visual features and tracking objects across video frames; (ii) \textbf{physics understanding and physics knowledge base}: reasoning on the extracted features (what it \emph{sees}) and retrieving representative physics laws from pretraining (what it \emph{knows}) to identify the underlying laws and estimate physical-parameter values; and (iii) \textbf{programming expertise}: synthesizing simulation code that handles boundary conditions and numerical-stability constraints, and regenerates physically and perceptually \textit{identical}, \textit{realistic} videos. Parameter estimation is a long-studied inverse problem traditionally addressed by physics-informed neural networks~\citep{raissi2019pinn}; we test whether MLLMs can solve it zero-shot~\citep{kojima2022zerocot}. This is not a retrieval or memorization task: every sample in our dataset uses randomly sampled parameters drawn from physically plausible ranges, so the model must infer numerical values from visual dynamics.

How good are the best MLLMs at this challenging inverse problem? Our evaluation shows that only a fraction of the generated code is executable or produces a video, and even when a video is generated the object interactions are often physically implausible or differ significantly from the input. Because our method explicitly probes the model's reasoning, we further observe that only a small fraction of cases correctly identify the underlying physical law and an even smaller fraction estimate the parameters within reasonable tolerance. While MLLMs have been extensively evaluated on inverse code generation for scientific \emph{images}, comparatively little work addresses scientific \emph{videos}, and this lack of comprehensive benchmarks has hindered progress in physically plausible simulation-video-to-code generation. Motivated by this gap, we construct \model: a benchmark of 160.6K samples in which each sample is a tuple of a physics-simulation video, the corresponding simulation code, a list of physical parameters and their values, and a detailed chain-of-thought (CoT) explanation~\citep{wei2022cot} describing the underlying physics using any valid formulation (including mixed Newtonian and Lagrangian approaches).

\noindent Our key contributions are:

\noindent \textbf{(1) \model}: the first comprehensive benchmark that combines physics-simulation video, executable code, parameter values, and physics explanations, making it possible to evaluate law identification, parameter recovery, and runnable simulation synthesis from video in a single framework.

\noindent \textbf{(2) Benchmarking insights}: extensive evaluation across a diverse set of closed- and open-source SOTA MLLMs, revealing their strengths and weaknesses across each aspect of the task.

\noindent \textbf{(3) Comprehensive evaluation strategy}: a holistic protocol combining verifiable metrics, LLM-based judges, and human assessment to measure different capabilities of MLLMs.

\section{Related Works}
\label{lit_review}

\begin{table*}[t]
\centering
\footnotesize
\setlength{\tabcolsep}{0.4em}
\caption{\textbf{Positioning against related physics and video benchmarks.} \textbf{Video}: evaluation grounded in video modality; \textbf{Text}: natural-language prompts; \textbf{CoT}: reasoning or chain-of-thought annotations; \textbf{Code}: executable simulation code as model output or dataset ground truth; \textbf{Physics Laws}: explicit governing laws in annotations or scoring; \textbf{Engine Agnostic}: the core benchmark pipeline spans multiple distinct simulation paradigms (such as analytic ODE-based and rigid-body-engine-based); \textbf{2D and 3D}: the benchmark spans both 2D and 3D visual content; \textbf{\# Exp.}: number of phenomena or experiment types; \textbf{Size}: total count of entries or videos reported in the source paper.}
\label{tab:benchmark_compare}
\begin{adjustbox}{width=\linewidth}
\begin{tabular}{lccccccccc}
\toprule
 & \multicolumn{4}{c}{\textbf{Grounding signals}}
 & \multicolumn{3}{c}{\textbf{Dataset properties}}
 & \multicolumn{2}{c}{\textbf{Dataset statistics}} \\
\cmidrule(lr){2-5} \cmidrule(lr){6-8} \cmidrule(lr){9-10}
\textbf{Benchmark}
 & \textbf{Video}
 & \textbf{Text}
 & \textbf{CoT}
 & \textbf{Code}
 & \textbf{Physics laws}
 & \textbf{Engine agnostic}
 & \textbf{2D+3D}
 & \textbf{\# Exp.}
 & \textbf{Size} \\
\midrule
Physion~\citep{bear2021physion} & \cmark & \xmark & \xmark & \xmark & \xmark & \xmark & \xmark & 8 & ${\sim}24$K \\
IntPhys~\citep{riochet2018intphys} & \cmark & \xmark & \xmark & \xmark & \xmark & \xmark & \xmark & -- & 15K \\
CLEVRER~\citep{yi2020clevrer} & \cmark & \cmark & \xmark & \xmark & \xmark & \xmark & \xmark & -- & 20K \\
ComPhy~\citep{chen2022comphy} & \cmark & \cmark & \xmark & \xmark & \xmark & \xmark & \xmark & -- & ${\sim}100$K \\
ContPhy~\citep{zheng2024contphy} & \cmark & \cmark & \xmark & \xmark & \xmark & \xmark & \xmark & -- & 6.5K \\
PHYBench~\citep{qiu2024phybench} & \xmark & \cmark & \xmark & \xmark & \cmark & \xmark & \xmark & -- & 500 \\
PhysBench~\citep{chow2025physbench} & \cmark & \cmark & \xmark & \xmark & \xmark & \xmark & \xmark & -- & 10{,}002 \\
VideoPhy~\citep{bansal2024videophy} & \xmark & \cmark & \xmark & \xmark & \xmark & \xmark & \xmark & -- & 11{,}330 \\
VBench~\citep{huang2024vbench} & \cmark & \cmark & \xmark & \xmark & \xmark & \xmark & \xmark & 8 & ${\geq}800$ \\
Morpheus~\citep{zhang2025morpheus} & \cmark & \cmark & \xmark & \xmark & \cmark & \xmark & \xmark & 9 & 130 \\
PhysInOne~\citep{zhou2026physinone} & \cmark & \cmark & \xmark & \xmark & \xmark & \cmark & \xmark & 71 & 2M \\
PhysCodeBench~\citep{xie2026physcodebench} & \xmark & \cmark & \xmark & \cmark & \cmark & \xmark & \xmark & -- & 700 \\
SimuScene~\citep{wang2026trainingbenchmarkingcodegeneration} & \xmark & \cmark & \cmark & \cmark & \cmark & \xmark & \xmark & 52 & 7659 \\
PhysVid~\citep{Zhan_2026_CVPR} & \cmark & \xmark & \xmark & \xmark & \xmark & \xmark & \xmark & -- & 12.1K \\
\rowcolor{blue!10} 
\textbf{\method (Ours)} & \cmark & \cmark & \cmark & \cmark & \cmark & \cmark & \cmark & 162 & 160.6K \\
\bottomrule
\end{tabular}
\end{adjustbox}
\end{table*}

We review the closest lines of work and identify the gap addressed by our video-to-simulation-code benchmark. Additional related-work details are provided in Appendix~\ref{app:additional_related_work}.

\textbf{Physical Law Recovery from Visual Dynamics:} Prior work has studied physical property estimation, law discovery, and visual intuitive physics from images and videos. Early systems such as Galileo~\citep{wu2015galileo} and Physics 101~\citep{wu2016physics101} combined physics engines with learning-based perception to infer object properties from video, while later work explored block-tower stability~\citep{lerer2016tower}, graph-based dynamics learning~\citep{battaglia2016interaction,watters2017vin}, and inverse-graphics formulations for recovering physical parameters~\citep{jaques2020pig,zhao2025toward}. Visual Physics~\citep{chari2019visualphysics} more directly targets physical law discovery from video trajectories. However, these approaches typically recover latent physical properties, equations, trajectories, or implicit per-particle dynamics, rather than producing a self-contained executable program from raw video. In contrast, our benchmark evaluates whether MLLMs can infer governing physical concepts and parameters from pixel-level video and also synthesize runnable simulation code that reproduces the observed dynamics.

\textbf{Benchmarks and Simulators for Physical Reasoning:} A complementary line of work benchmarks physical understanding in generative and reasoning models. PhysInOne~\citep{zhou2026physinone}, VideoPhy~\citep{bansal2024videophy}, Morpheus~\citep{zhang2025morpheus}, VBench~\citep{huang2024vbench}, PhysBench~\citep{chow2025physbench}, PhyBench~\citep{qiu2024phybench}, Physion~\citep{bear2021physion}, IntPhys~\citep{riochet2018intphys}, CLEVRER~\citep{yi2020clevrer}, and ComPhy~\citep{chen2022comphy} evaluate physical plausibility, commonsense reasoning, invariant preservation, property estimation, or future-state prediction. These benchmarks probe whether models recognize, judge, or predict physical behavior, but they do not require recovering an executable simulation whose trajectory can be compared against the original video at the level of laws, parameters, and code. Related work also uses simulators for physics reasoning and learning, like simulator-generated QA for physics problem solving~\citep{prabhudesai2026physolym}, and the popular simulators are MuJoCo~\citep{todorov2012mujoco}, DiffTaichi~\citep{hu2020difftaichi}, Brax~\citep{freeman2021brax}, Genesis~\citep{genesis2024}, Isaac Gym~\citep{makoviychuk2021isaac}, and NeuralSim~\citep{heiden2021neuralsim}. The closest benchmark, PhysCodeBench~\citep{xie2026physcodebench}, generates code for a single physics engine from text inputs, whereas ours spans multiple engines, uses standard Python with engine-specific starter code (testing physics understanding over engine familiarity), evaluates MLLMs' visual capabilities from video inputs, and is substantially larger (160.6K vs.\ 700 samples) with extensive regenerated-video-based physics evaluation that PhysCodeBench lacks. Our work thus complements prior benchmarks by focusing on scalable, visually grounded physics-understanding evaluation.

\section{Dataset Overview}
\label{dataset}
\subsection{Dataset Curation Process}
\label{curation_pipeline}
\begin{figure}[t]
\centering
\includegraphics[width=\textwidth]{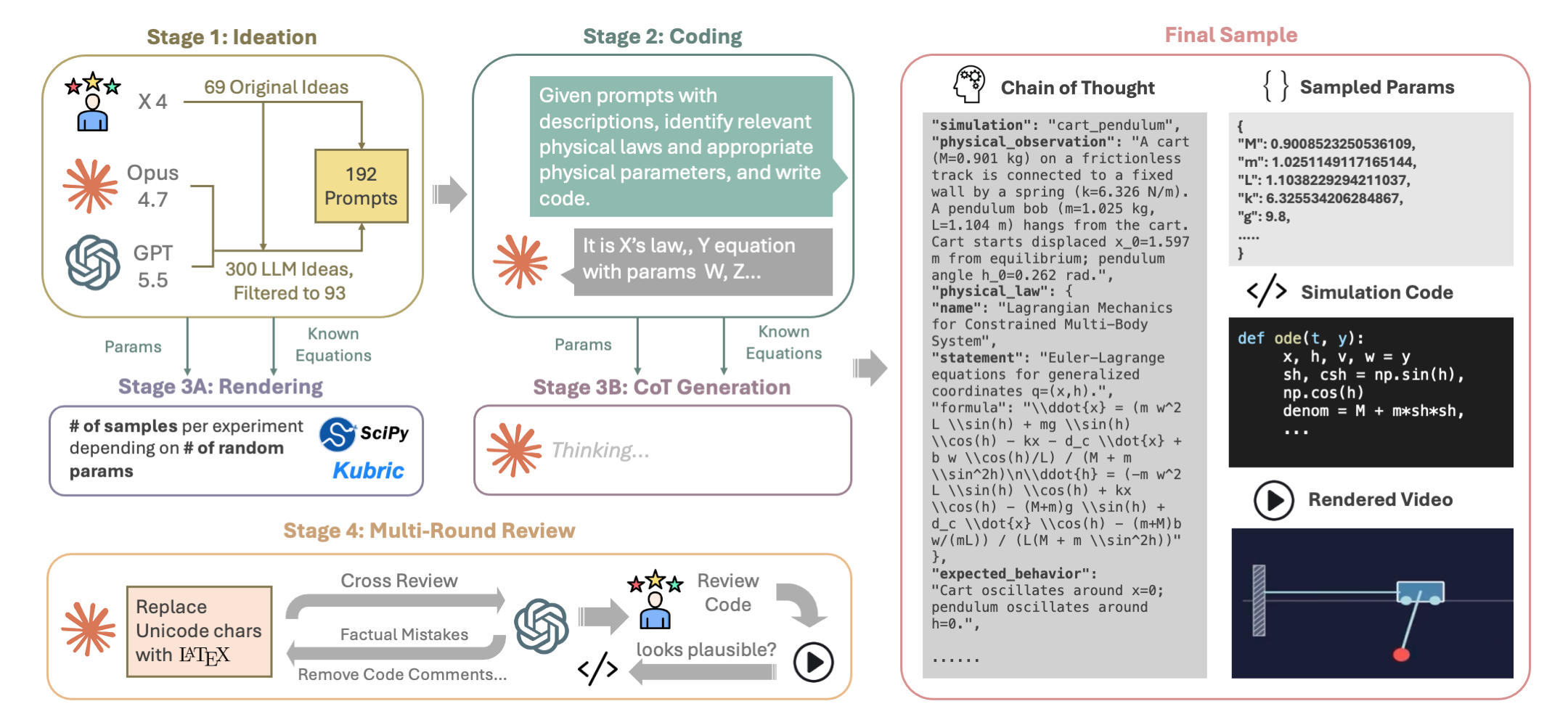}
\caption{The dataset curation pipeline.}
\vspace{-2mm}
\label{fig:dataset_curation}
\end{figure}

Dataset construction proceeds through two complementary pipelines that together cover 162 distinct physics phenomena (Figure~\ref{fig:dataset_curation}). Regardless of the pipeline, every sample comprises a Python simulation script, the rendered simulation video, a structured CoT annotation, and the underlying parameter values. We use Python as the common implementation language; it is widely represented in LLM training data, so models tend to generate it well, relying on solver libraries such as SciPy~\citep{2020SciPy-NMeth} or physics engines accessible through Python wrappers such as PyBullet~\citep{coumans2016pybullet}. The pipeline is not engine-specific: it can be extended to any physics engine with a Python wrapper and can generate an indefinite number of samples by randomizing all physical parameters.

Four human experts first curated 69 physics-simulation ideas from their own knowledge, then prompted Claude Opus 4.7 and GPT 5.5 to propose additional candidates; from the 300 generated scenes the experts filtered and refined 93 based on implementability and clarity of the intended physics laws, yielding the 162 experiments, accompanied by 3--4 line event descriptions written entirely by the experts. These descriptions were used to prompt Claude Opus 4.7 to generate the Python simulation code, extract the relevant physics laws, and identify the physical parameters appearing in the equations from its pretrained knowledge. The generated code was executed and the resulting videos iteratively refined through 2--10 rounds of human-in-the-loop correction until physically plausible. Finally, the physics knowledge and sampled parameters were recorded in a structured CoT (Table~\ref{tab:cot_schema}), with parameter values drawn uniformly from physically plausible ranges to produce multiple samples per experiment.

\textbf{Pipeline A} produces 2D simulation data for 87 experiments entirely through standard Python libraries (SciPy, Math). Each module is structured around a parameterised dataclass, a numerical ODE solver, a renderer, and a code exporter, and is inspired by implementations from MyPhysicsLab\footnote{\url{https://www.myphysicslab.com/}}, a widely used educational physics-simulation library.

\textbf{Pipeline B} (denoted "Kubric" throughout, after the 3D rigid-body domain it targets, but implemented with PyBullet + pyrender + trimesh because the Kubric\footnote{\url{https://kubric.readthedocs.io/en/latest/}} package itself requires Blender + Docker) covers 75 3D experiments needing contact-aware articulated and rigid-body dynamics beyond analytic ODE formulations. Each experiment is a self-contained module built on a shared infrastructure that exposes a uniform API for primitive meshes, joints, constraints, ropes, and granular media, and is structured around a parameterised dataclass, a parameter sampler that draws physically plausible values per seed, and a scene constructor.

\textbf{Ground Truth Debiasing}: Because the ground-truth code and CoT were initially generated by Claude and may carry Claude-specific stylistic biases or factual errors, we ran a three-round LLM-assisted debiasing pipeline followed by human verification. Round~1: Claude Opus 4.7 removed Unicode symbols from equations and replaced them with LaTeX and Python forms that are more suitable for MLLM training. Rounds~2-3: GPT-5.5 corrected factual and physics-related errors, fixed inconsistencies between LaTeX and Python equations, and stripped code comments to eliminate Claude-specific coding-style bias during code-similarity evaluation, making $14{,}700$ substitutions for a total Levenshtein distance of $693{,}600$ characters. Finally, four human experts manually verified the CoTs and applied additional corrections, yielding a ground truth that is both reliable and debiased. To further ensure fair evaluation (Section~\ref{metrics}), we place limited emphasis on code similarity; many different implementations can produce the same simulation, and instead focus on simulation equivalence and the physical plausibility of the regenerated videos. We discuss more rigorous evaluations proving removal of Claude authorship bias in Section \ref{app:claude_bias}.

\subsection{Dataset Statistics}

\paragraph{Scale.} The 162 categories are unevenly split across the two pipelines because each pipeline targets a different physical regime, and the sample count per category scales with the number of independently randomized parameters. Pipeline~A's 87 SciPy categories are analytically formulated and inexpensive to integrate, so we allocate $100 \times n_{\text{random\_params}}$ samples per category, ranging from 200 (low-DOF systems such as \texttt{brachistochrone\_race}) to 2{,}500 for seed-driven experiments that have high latent dimensionality (\texttt{complex\_collision\_plane}: 2--6 randomly-typed objects with per-object mass, size, and initial conditions). Pipeline~B's 75 Kubric categories require GPU rendering and benefit from a wider visual diversity, so we apply the same $100 \times n_{\text{random\_params}}$ samples per category, then $+50\%$ if the scene has a variable object count, then $+30\%$ if it has color/material randomization, floored at 400 (e.g. \texttt{billiard\_break\_3d}), capped at 5,000 (e.g. \texttt{multi\_physics\_arena}). There are \textbf{64{,}759} SciPy samples (mean 744, median 600 per category) and \textbf{95{,}855} Kubric samples (mean 1{,}278, median 900 per category) for a dataset total of \textbf{160{,}614} video--code--CoT--parameter tuples (more statistical details in Section \ref{app_stat}). 

\begin{figure}[ht]
\centering
\includegraphics[width=\linewidth]{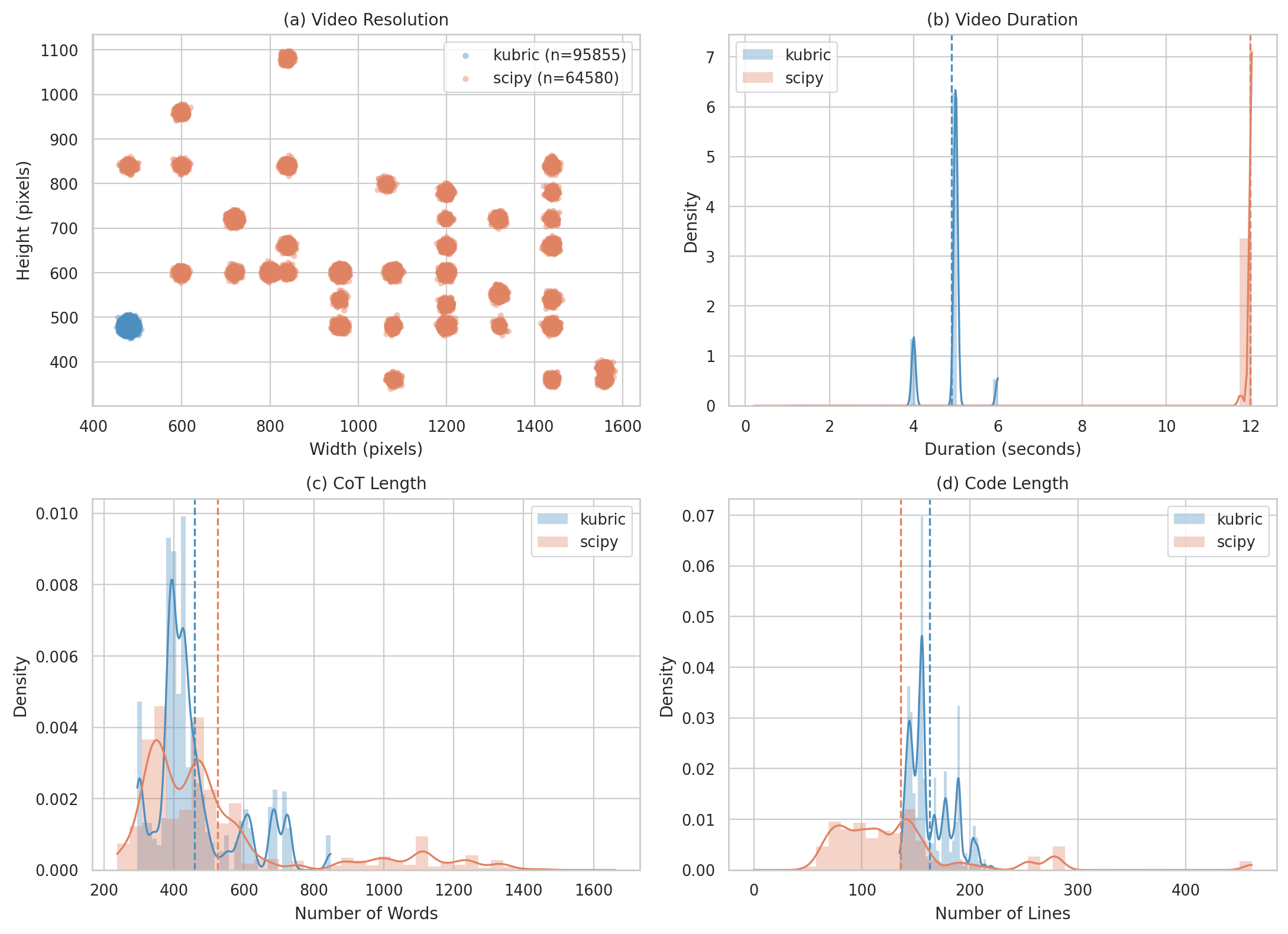}
\caption{Per-sample distributions of (a)~video resolution, (b)~duration,
(c)~CoT length, and (d)~code length across Kubric
and SciPy. Dashed lines mark per-pipeline means.}
\vspace{-2mm}
\label{fig:dataset_distributions}
\end{figure}

\paragraph{Distribution by Physical Domain and Law.} We cover diverse experiments across rigid mechanics, waves and oscillations, and some fluid and magnetic phenomena. Figure \ref{fig:pie} shows the counts of the 9 main categories of physical domains covered across the 162 experiments (more about these 9 physical domains in Section \ref{app_stat}) and some dataset examples from each category. Appendix Tables~\ref{tab:scipy_per_experiment} and~\ref{tab:kubric_per_experiment} list all experiments with their random-parameter and sample counts. Figure \ref{fig:dataset_distributions} shows that Kubric's 95{,}855 samples are fixed at 480$\times$480 over 5.0s median duration (range 1.1--6.0s) with tight CoTs (median 428 words, range 295--848) and uniformly templated code (median 156 lines, range 135--232), whereas SciPy's 64,759 samples span wider 480$\times$360 to 1560$\times$1080 resolutions and longer 12.0s median durations (range 0.2--12.0s) with longer-tailed CoTs (median 452 words, up to 1{,}662) and broader code lengths (median 119, range 0--461).

\section{Methodology}
\label{Methods}

\paragraph{Task definition.}
The task is to watch a video of a physical process and infer the underlying physical law and equations, estimate key parameters, and recreate a working simulation. Because an MLLM may not be well pretrained on the variety of physics engines available, we provide the engine name and an experiment-agnostic code template as a starting point, ensuring we evaluate the model's understanding of physics rather than its familiarity with a specific engine. Formally, given only a rendered video, a model must produce (i) a structured chain-of-thought JSON object following the engine-conditional 4-key schema, a subset of the ground-truth CoT keys (Table \ref{tab:cot_schema}), and (ii) executable Python code that simulates the observed process. 

\subsection{Evaluation Metrics}
\label{metrics}

We want to measure the three capabilities introduced in Section~\ref{intro}: video understanding, physics understanding, and programming abilities of MLLMs. We explain the correspondence of our proposed evaluation metrics with the measured capabilities in Table \ref{tab:capabilities}. Video understanding is the prerequisite capability that every downstream metric passes through. It is observable through video appearance similarity; a very low video-similarity score implies the model failed to read the scene at all. 

Physics understanding rests on (i) the model's prior knowledge of physical laws and (ii) its ability to infer the laws demonstrated in a given video; to disentangle the two, we (a) judge the \textbf{absolute} correctness of the stated laws regardless of the video to test the model's prior physics knowledge, and (b) compare the laws in each model's generated CoT against those in the dataset's reference CoT. Both are rated on a 1--5 Likert scale (5 = best match, 1 = lowest) averaged over three LLM judges: Gemini-2.5-flash, Grok-4.1-Fast, and GPT-5-Nano. Apart from evaluating the physics laws, we also verify parameter recovery. We extract the ground-truth physical parameters from the reference CoT (excluding scene metadata, geometry, color, and the gravitational constant) and the predicted parameters from each model's emitted CoT, pair predicted parameter names to ground-truth names by greedy bipartite matching over Jaccard similarity on alias sets that capture lexical and Greek-letter abbreviations, normalize units, and compute the per-pair relative error $|\hat{x}-x|/|x|$ (absolute error as fallback when $|x|$ is below numerical-zero tolerance); a parameter is counted as correctly estimated if its relative error is within $\pm 20\%$. 

For coding and instruction following we compute verifiable metrics: code executability, syntax correctness, adherence to the CoT-then-code output format, and whether the code successfully writes the video MP4, together with CodeBLEU against the reference code. Finally, we measure end-to-end capabilities from the rendered videos by executing the generated code through (i) standalone physical plausibility, (ii) physics equivalence (same physical phenomenon as the ground truth). For each model, four human experts independently rate the same 100 random samples on (i) physical plausibility, (ii) physics equivalence, and (iii) video appearance similarity (same object shape, count, color, etc.); we report the mean over the four raters and inter-rater Cohen's $\kappa$ in Figure~\ref{fig:four_images}. Metric (iii) reveals whether models lean toward replicating appearance over physics. For scalability, we also report appearance similarity directly via the patch-wise video-similarity metrics DINOv2 and VideoCLIP. In Table \ref{tab:capabilities}, stages (i) and (iii) are largely solved; stage (ii) is the bottleneck, and that is the claim \model is built to isolate. 

\begin{table}[t]
\centering
\caption{How different stages of capabilities of MLLMs are measured through all evaluation metrics.}
\label{tab:capabilities}
\begin{tabular}{ll}
\toprule
\textbf{Stage} & \textbf{Evaluation / Measurement} \\
\midrule
(i) Video understanding
& Appearance equivalence (human), DINOv2/VideoCLIP (Table~\ref{tab:video-sim-equivalence}) \\
(ii) Physics understanding
& Law correctness \& equivalence (Table~\ref{tab:likert_scores}); parameter recovery (Tables~\ref{tab:code_generation_metrics}, \ref{tab:param-eval-detailed}) \\
(iii) Programming
& Compilation; execution; video generation; CodeBLEU (Table~\ref{tab:code_generation_metrics}) \\
\textit{End-to-end}
& \textit{Physical plausibility \& equivalence (Table~\ref{tab:video-sim-equivalence})} \\
\bottomrule
\end{tabular}
\end{table}

\subsection{Benchmarking Details}
\label{benchmarking}

Our generated dataset contains 160{,}614 samples spanning 162 distinct physics experiments (87 SciPy, 75 Kubric); within an experiment, samples differ only in random initial conditions and parameter values, while the underlying physical concept and required code structure remain fixed. We therefore treat each experiment as a natural unit of analysis and evaluate every model on $n=15$ uniformly random samples per experiment under a single seed ($s=42$), yielding $162\times15=2{,}430$ evaluations per model, a $\sim$66$\times$ reduction over the full corpus that we accept on statistical grounds (see Section \ref{app:stat}), so the relative rankings established on the 2,430-sample subset are statistically supported and very likely to hold on the full 160,614-sample corpus, with reduced API cost.

\textbf{Models evaluated}: On our 2{,}430-sample benchmarking dataset, we benchmark 10 MLLMs, five closed-source models: Claude Sonnet~4.6, GPT-5-mini, Grok-4-fast-reasoning, Gemini~2.5~Pro, and Amazon Nova 2 Lite,
and five open-source models: InternVL3.5-30B-A3B, Qwen3-VL-30B-A3B, Pixtral-12B,
Gemma3-12B-Instruct, and GLM-4.1V-9B-Thinking. All 10 models receive the
\emph{same} engine-conditioned prompt (Appendix Section \ref{app:prompt}). More benchmarking results with GPT-6-Astra, Claude Fable-5.1, Kimi K3, and Grok-4.6 on 10 handpicked hard samples are in Section~\ref{sota}.

\textbf{Input format and decoding}: For frame-based models (Claude, GPT-5-mini, Grok-4, and all five open-source models), we decode every frame at 10\,FPS into a list of JPEG-encoded $360{\times}180$ (\texttt{scipy}) or $240{\times}240$
(\texttt{kubric}) RGB images and pass them as multi-image messages alongside the engine name. Gemini~2.5~Pro instead consumes the raw \texttt{.mp4} at the same FPS and resolution as others, since it natively supports video. We use a context length of $32{,}768$ tokens for the open-source models (49{,}152 for Pixtral, which has
larger per-frame token cost), greedy decoding where supported, and per-model output-token budgets are 8K (Qwen3-VL, Pixtral, InternVl-3.5, Gemma-3), 9K (GPT, Claude, Nova), 12K (GLM-Thinking, Grok), and 16K (Gemini, to accommodate hidden-thinking tokens). Closed-source inference runs at concurrency 200 against the providers' endpoints, open-source inference runs as eight parallel vLLM shards on $8{\times}$NVIDIA H200 (144\,GB HBM each) with \texttt{bfloat16} compute and prefix caching enabled.

% --- Likert Scores ---
\begin{table}[t]
\centering
\footnotesize
\setlength{\tabcolsep}{1em}
\caption{\textbf{Average Likert scores} (on a 1--5 scale) rated by 3 LLM judge models (Gemini-2.5-flash, Grok-4.1-Fast, GPT-5-Nano) for model outputs across physical law correctness and equivalence.}
\label{tab:likert_scores}
\begin{adjustbox}{width=\linewidth}
\begin{tabular}{l|cc|cccc}
\toprule
& \multicolumn{2}{c|}{\textbf{Physical Law Correctness}} & \multicolumn{4}{c}{\textbf{Physical Law Equivalence}} \\
\textbf{Model} & \textbf{Formula} & \textbf{Overall} & \textbf{Name} & \textbf{Statement} & \textbf{Formula} & \textbf{Overall} \\
\midrule
\rowcolor{gray!15} \multicolumn{7}{l}{\textit{Open-source models}} \\
\textsc{Qwen3-VL-30B-A3B-Instruct} & 4.58 & 4.68 & \textbf{2.18} & \textbf{2.09} & \textbf{1.92} & \textbf{2.02} \\
\textsc{InternVL3.5-30B-A3B}       & 4.68 & 4.64 & 1.98 & 1.96 & 1.75 & 1.82 \\
\textsc{Gemma-3-12B-IT}            & \textbf{4.90} & \textbf{4.93} & 2.15 & 1.95 & 1.72 & 1.90 \\
\textsc{GLM-4.1V-9B-Thinking}      & 4.46 & 4.43 & 2.14 & 2.03 & 1.80 & 1.91 \\
\textsc{Pixtral-12B-2409}          & 4.43 & 4.50 & 1.70 & 1.57 & 1.45 & 1.51 \\
\midrule
\rowcolor{gray!15} \multicolumn{7}{l}{\textit{Closed-source models}} \\
\textsc{GPT-5 Mini}                & 4.75 & \textbf{4.70} & \textbf{2.71} & \textbf{2.69} & \textbf{2.24} & \textbf{2.47} \\
\textsc{Gemini 2.5 Pro}            & \textbf{4.76} & 4.66 & 2.49 & 2.48 & 2.16 & 2.34 \\
\textsc{Claude Sonnet 4.6}         & 4.53 & 4.33 & 2.58 & 2.45 & 2.09 & 2.26 \\
\textsc{Grok 4 Fast Reasoning}     & 4.65 & 4.58 & 2.16 & 2.16 & 1.85 & 1.98 \\
\textsc{Nova 2 Lite}               & 4.29 & 4.27 & 1.80 & 1.87 & 1.59 & 1.69 \\
\bottomrule
\end{tabular}
\end{adjustbox}
\end{table}

%----Human evaluation ----
\begin{table}[t]
\centering
\caption{\textbf{Video simulation equivalence table} containing human judgment of Physical
Plausibility, Video's Physics Equivalence, Appearance Equivalence, and verifiable video frame
similarity scores using DINOv2 and VideoCLIP. Human scores on 1--5 Likert (on per model 100
random samples).}
\label{tab:video-sim-equivalence}
\begin{adjustbox}{width=\linewidth}
\footnotesize
\setlength{\tabcolsep}{0.4em}
\begin{tabular}{l|ccc|cc}
\toprule
 & \multicolumn{3}{c|}{\textbf{Human Evaluation}} & \multicolumn{2}{c}{\textbf{Patchwise Video Similarity}} \\
\cmidrule(lr){2-4} \cmidrule(lr){5-6}
\textbf{Model} & \textbf{Physical Plausibility} & \textbf{Physics Equivalence} & \textbf{Appearance Equivalence} & \textbf{DINOv2 sim.} & \textbf{VideoCLIP sim.} \\
\midrule
\rowcolor{gray!15} \multicolumn{6}{l}{\textit{Open-source models}} \\
\textsc{Qwen3-VL-30B-A3B-Instruct} & \textbf{3.16} & \textbf{1.88} & 2.31           & \textbf{0.557} & 0.713          \\
\textsc{InternVL3.5-30B-A3B}       & 1.43          & 1.23          & 1.59           & 0.482          & 0.667          \\
\textsc{Gemma-3-12B-IT}            & 2.64          & 1.80          & \textbf{2.56}  & 0.447          & 0.643          \\
\textsc{GLM-4.1V-9B-Thinking}      & 2.43          & 1.32          & 2.04           & 0.551          & \textbf{0.732} \\
\textsc{Pixtral-12B-2409}          & 1.48          & 1.13          & 1.26           & 0.448          & 0.638          \\
\midrule
\rowcolor{gray!15} \multicolumn{6}{l}{\textit{Closed-source models}} \\
\textsc{GPT-5 Mini}                & 3.40          & 2.40          & 2.66           & 0.672          & 0.785          \\
\textsc{Gemini 2.5 Pro}            & \textbf{3.96} & \textbf{2.86} & \textbf{3.45}  & \textbf{0.774} & \textbf{0.838} \\
\textsc{Claude Sonnet 4.6}         & 3.30          & 2.30          & 3.14           & 0.767          & 0.837          \\
\textsc{Grok 4 Fast Reasoning}     & 1.90          & 1.65          & 2.28           & 0.629          & 0.750          \\
\textsc{Nova 2 Lite}               & 2.62          & 1.14          & 1.63           & 0.480          & 0.687          \\
\bottomrule
\end{tabular}
\end{adjustbox}
\end{table}

% --- Code Generation ---
\begin{table*}[t]
\centering
\footnotesize
\setlength{\tabcolsep}{0.8em}
\caption{\textbf{Evaluation of parameter estimation accuracy and generated simulation code quality}. Percentages computed as a fraction of all N=2430 inputs that successfully reached each stage (CoT extracted, Code extracted, Compiles, Runs without error, Saves video).}
\label{tab:code_generation_metrics}
\begin{adjustbox}{width=\linewidth}
\begin{tabular}{l|c|cccccc}
\toprule
& \textbf{Param. Estimation} & \multicolumn{6}{c}{\textbf{Code Generation and Execution}} \\
\textbf{Model} & \textbf{Acc. (\(\pm20\%\))} & \textbf{CoT} & \textbf{Code} & \textbf{Compile} & \textbf{No Errors} & \textbf{Video Save} & \textbf{CodeBLEU} \\
\midrule
\rowcolor{gray!15} \multicolumn{8}{l}{\textit{Open-source models}} \\
\textsc{Qwen3-VL-30B-A3B-Instruct} & \textbf{6.16\%} & 99.8\% & 96.8\% & 96.6\% & 73.8\% & 77.9\% & 0.206 \\
\textsc{InternVL3.5-30B-A3B}       & 3.52\% & 99.8\% & 99.7\% & 94.0\% & 48.0\% & 62.5\% & 0.195 \\
\textsc{Gemma-3-12B-IT}            & 3.98\% & 99.0\% & 95.4\% & 95.2\% & 39.6\% & 47.7\% & 0.200 \\
\textsc{GLM-4.1V-9B-Thinking}      & 4.84\% & 96.6\% & 87.6\% & 87.4\% & 46.6\% & 48.0\% & 0.198 \\
\textsc{Pixtral-12B-2409}          & 3.31\% & 99.8\% & 99.7\% & 99.0\% & 42.6\% & 60.7\% & 0.187 \\
\midrule
\rowcolor{gray!15} \multicolumn{8}{l}{\textit{Closed-source models}} \\
\textsc{GPT-5 Mini}                & 7.13\% & 83.7\% & 98.6\% & 98.2\% & 92.8\% & 93.3\% & 0.241 \\
\textsc{Gemini 2.5 Pro}            & 6.06\% & 96.1\% & 66.2\% & 66.1\% & 58.7\% & 59.2\% & 0.223 \\
\textsc{Claude Sonnet 4.6}         & \textbf{7.69\%} & 100.0\% & 99.9\% & 99.8\% & 98.8\% & 99.0\% & 0.232 \\
\textsc{Grok 4 Fast Reasoning}     & 5.69\% & 98.9\% & 100.0\% & 99.6\% & 89.4\% & 90.4\% & 0.209 \\
\textsc{Nova 2 Lite}               & 3.76\% & 97.0\% & 99.0\% & 98.5\% & 75.9\% & 78.7\% & 0.204 \\
\bottomrule
\end{tabular}
\end{adjustbox}
\end{table*}

\section{Results and Discussions}
\label{results}

We evaluate models along four axes: (i) extracted physical-law correctness and match with ground truth (Table~\ref{tab:likert_scores}); (ii) video-simulation equivalence by human judgment and verifiable patch-wise similarity (Table~\ref{tab:video-sim-equivalence}); (iii) parameter-estimation correctness; and (iv) code correctness and similarity (both Table~\ref{tab:code_generation_metrics}). Claude leads only on parameter estimation (with GPT-5-mini close), suggesting our code-style debiasing pipeline removes the most direct Claude-specific advantage (coding).

\textbf{Physical Law Understanding}: Both open- and closed-source models struggle to infer physics from video (Table~\ref{tab:likert_scores}). Most models (7 of 10) score $>$4.5/5 on stating physical laws and formulae, indicating decent \emph{prior} physics knowledge, but this fails to translate into recognizing the laws actually demonstrated in a video, the best model, GPT-5 Mini, scores only 2.47/5, and closed-source systems average just 2.15 vs.\ 1.83 for open-source on this harder task. 

\begin{figure*}[t]
\centering
\begin{subfigure}[t]{0.40\linewidth}
    \centering
    \includegraphics[width=\linewidth]{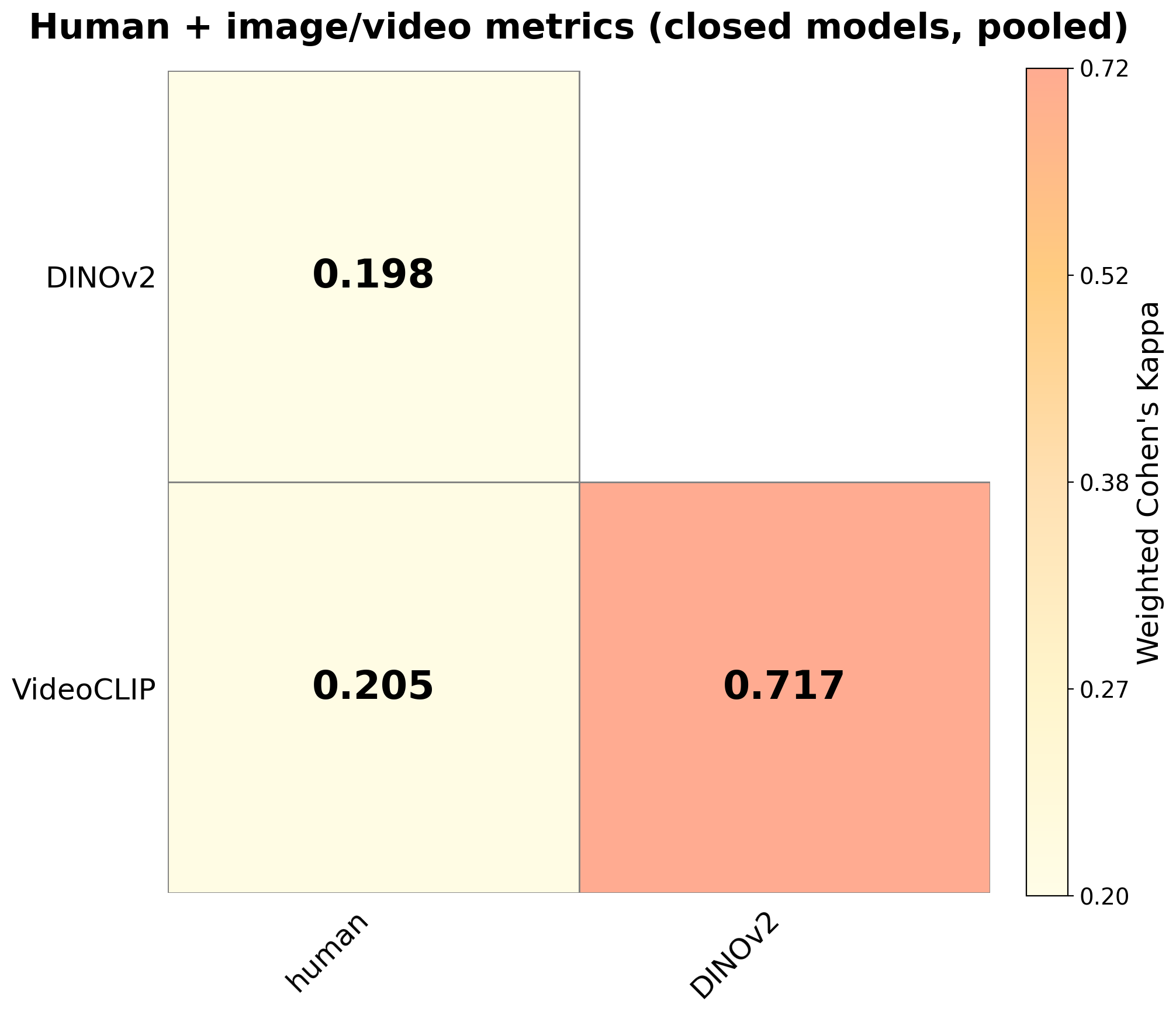}
    \caption{}
    \label{fig:a}
\end{subfigure}
\hfill
\begin{subfigure}[t]{0.40\linewidth}
    \centering
    \includegraphics[width=\linewidth]{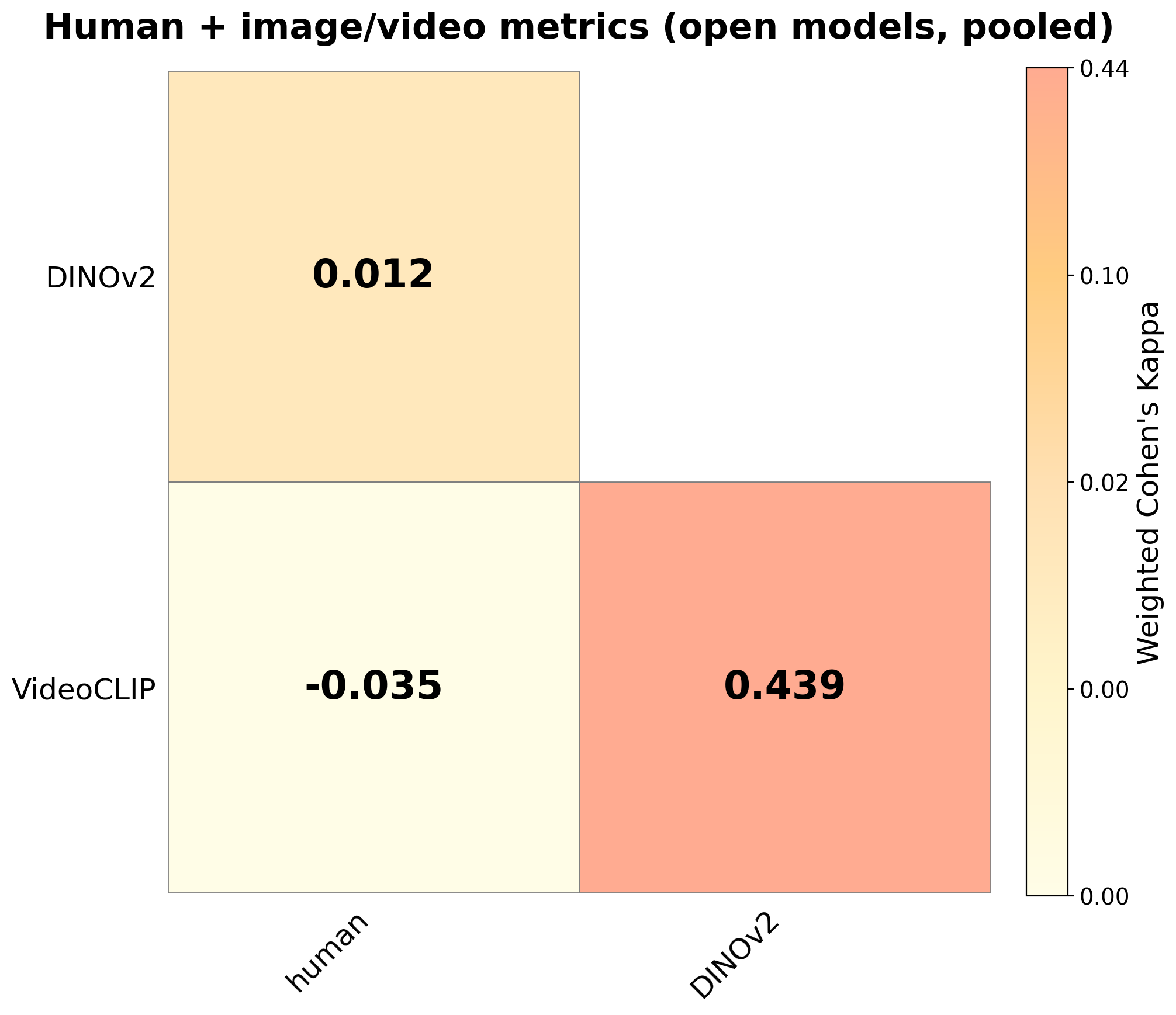}
    \caption{}
    \label{fig:b}
\end{subfigure}

\par\medskip

% Second row: (c) and (d)
\begin{subfigure}[t]{0.40\linewidth}
    \centering
    \includegraphics[width=\linewidth]{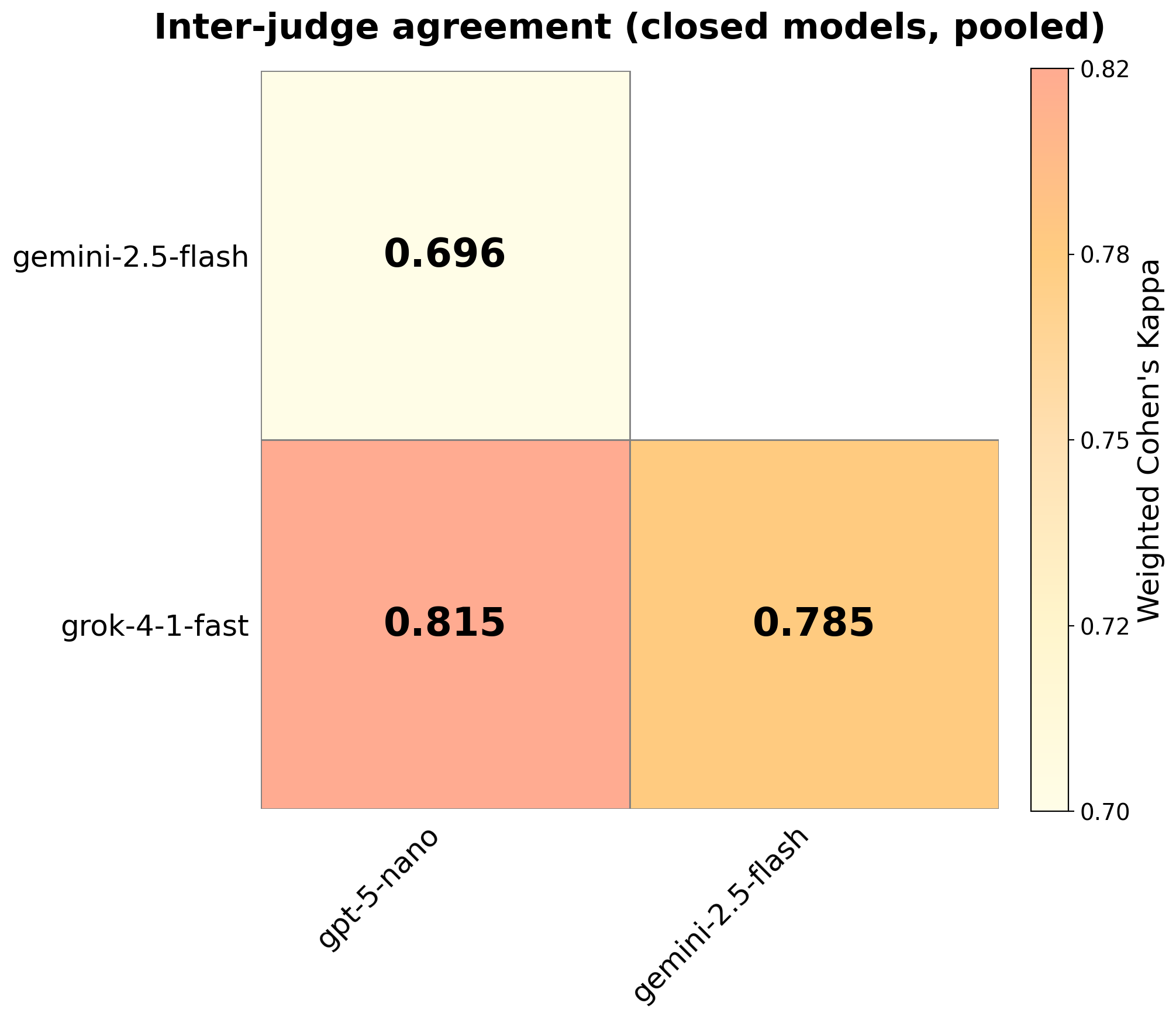}
    \caption{}
    \label{fig:c}
\end{subfigure}
\hfill
\begin{subfigure}[t]{0.40\linewidth}
    \centering
    \includegraphics[width=\linewidth]{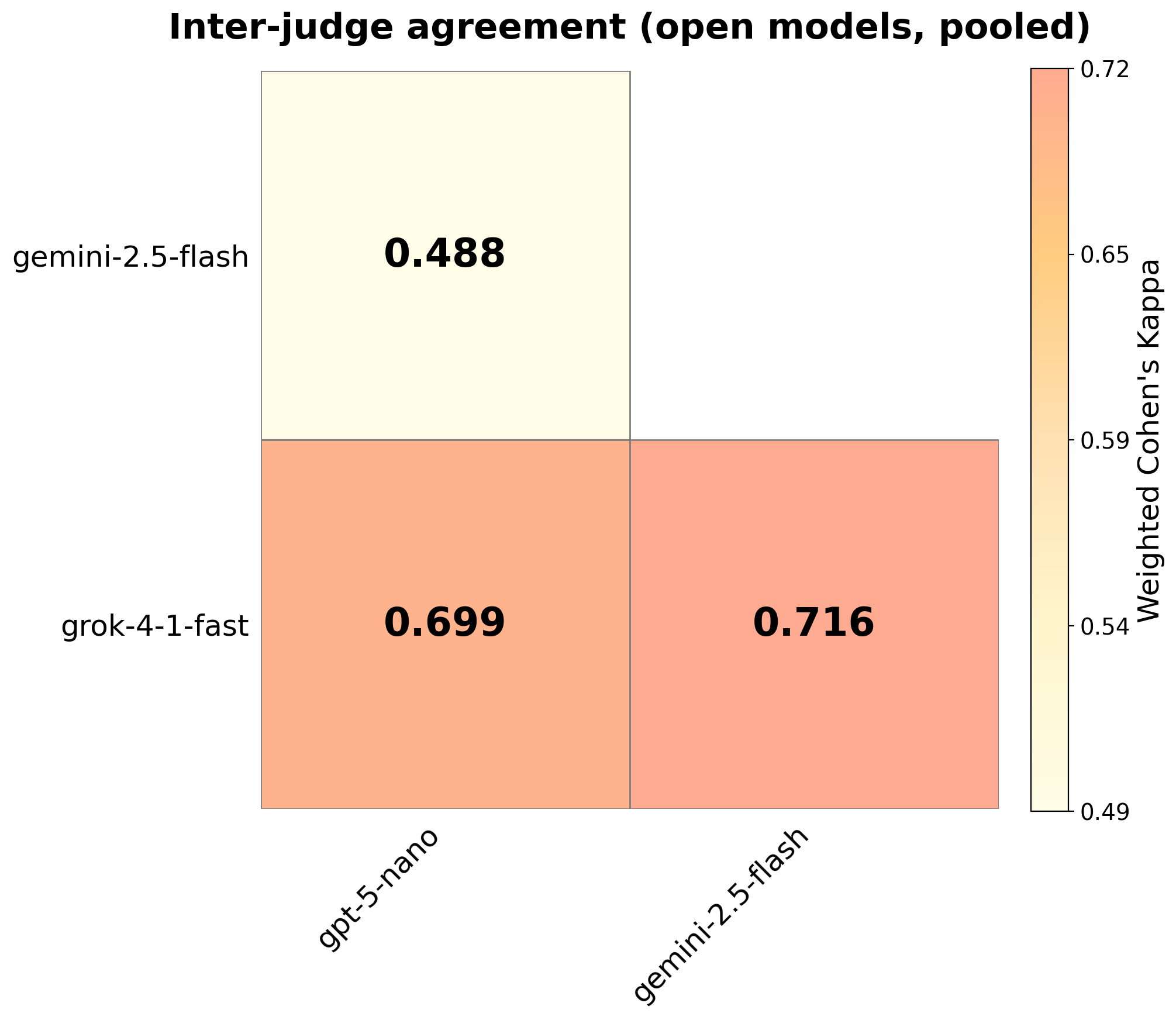}
    \caption{}
    \label{fig:d}
\end{subfigure}

\caption{Cohen's $\kappa$ for alignment between human-rated physics equivalence of videos and the patch-wise similarity metrics (DINOv2, VideoCLIP), averaged over (a) closed- and (b) open-source models, and inter-judge Cohen's $\kappa$ averaged over (c) closed- and (d) open-source models.}
\label{fig:four_images}
\end{figure*}

\textbf{Simulation Equivalence}: Table~\ref{tab:video-sim-equivalence} reveals a consistent open-vs.-closed gap on video-simulation equivalence (human rating guidelines in Section \ref{app:guideline}), with closed models stronger on both human judgment and embedding-based similarity. \textsc{Gemini 2.5 Pro} ranks highest overall, leading on human-rated Physical Plausibility, Physics Equivalence, and Appearance Equivalence as well as DINOv2 and VideoCLIP similarity; despite this, Table~\ref{tab:code_generation_metrics} shows Gemini lagging other closed models on code executability, which we attribute to its long thinking trace consuming most output tokens and truncating the code. We observe that across all models the patchwise video similarity metrics are far more lenient than human evaluations of physics equivalence or appearance equivalence of videos since these patchwise metrics are driven by pixel-level semantics, while humans are very sensitive to any deviation from reality. Across all models, multimodal systems are considerably better at preserving visual appearance than at modeling or verifying physical behavior: Figure~\ref{fig:four_images} shows that DINOv2 and VideoCLIP correlate highly with each other but not with human raters, who judge physics equivalence (whether the regenerated video exhibits the same physics laws as the GT); the figure also shows the maximum pairwise correlation between any two judges is $<0.85$, so our judge set is not highly correlated and is reliable (model-specific bias is eliminated).

\textbf{Parameter Estimation Correctness}: Table~\ref{tab:code_generation_metrics} reports the fraction of all ground-truth physical parameters each model both names correctly \emph{and} estimates within $\pm 20\%$, symmetrically penalizing naming failures and value errors. Appendix Table~\ref{tab:param-eval-detailed} decomposes this into naming recall and value accuracy: across the 10 MLLMs, \textsc{Claude-Sonnet-4.6} and \textsc{GPT-5-Mini} are strongest at naming the right parameters, while \textsc{GLM-Thinking} achieves the highest per-pair accuracy on the (small) set of parameters it actually names. Claude names $70\%$ more parameters per sample at near-equal per-pair accuracy ($18.5\%$), yet recovers only $7.7\%$ of all ground-truth parameters within tolerance, indicating that quantitative physics-from-video estimation is substantially harder than qualitative scene description.

\textbf{Code Correctness}: Table~\ref{tab:code_generation_metrics} shows that almost every model generates the CoT and syntactically correct code for almost all samples, the only exception being Gemini, which fails to emit code in $>$30\% of samples. However, open-source code often does not execute: averaged across the five open models only $50.1\%$ runs to completion without errors and $59.4\%$ writes a video file, versus an average closed-source success rate of $83.1\%$, with \textsc{Claude Sonnet 4.6} the most reliable overall ($98.8\%$ error-free execution, $99.0\%$ video-write). CodeBLEU is similar across models and we caution it is partially unreliable: the same physics phenomenon admits different code.

\section{Conclusion}
\label{conclusion}
We presented \model, a procedurally generated benchmark of 160.6K (video, simulation code, chain-of-thought, parameters) tuples spanning 162 experiments across two simulation engines, paired with an LLM-judge, human-rater, and verifiable scoring suite for holistic MLLM physics-understanding evaluation. Evaluating ten frontier MLLMs reveals four consistent gaps: models can state physics laws (most score $>$4.5/5 Likert) but struggle to infer them from video (best 2.47/5, GPT-5 Mini); closed-source models dominate qualitative simulation equivalence, with Gemini 2.5 Pro leading visual-physics ratings and DINOv2/VideoCLIP similarity and Claude Sonnet 4.6 leading code-execution reliability (98.8\%). End-to-end parameter recovery is severely limited, with the best model (Claude Sonnet 4.6) recovering only $\sim$7.7\% of ground-truth parameters within $\pm 20\%$; and video appearance reproduction is far easier than physics reproduction. Two limitations remain: (i) rigid-body dynamics dominate the benchmark, leaving only a small fraction of fluid/deformable scenarios that an MPM-based engine such as Genesis would naturally extend and (ii) these benchmark videos are simulator-generated, so success on this benchmark doesn't prove success in real-world noisy videos. However, as discussed in \ref{app:transfer2real}, we believe that our benchmark reports failure in the absence of nuisance factors isolating the cause to MLLM capabilities. We hope the released evaluation set and evaluation pipeline accelerate community progress on physics-grounded video understanding.

\newpage

\newpage

\subsubsection*{Acknowledgments}
We acknowledge the contribution of Chujun Tang, Brown University, in helping us to create some of the figures of this paper. This work was financially supported by Amazon. Yizhou Zhao was supported in part by the SoftBank Group–ARM Fellowship.

\bibliography{references}
\bibliographystyle{preprint}

\appendix
\section{Appendix}

\subsection{Evaluating Latest Frontier Models}
\label{sota}
As the latest LLMs are solving previously unprecedented multimodal tasks better, we wanted to test our benchmark (which turned out to be hard for the models evaluated in the main paper) on the latest frontier models. Due to cost concerns, we evaluate four frontier models on a handpicked set of 10 hard samples from 5 SciPy experiments and 5 Kubric experiments to get a sense of how the latest frontier models stand against the 10 models evaluated in our benchmark. We test three closed-source frontier models: Claude Fable 5.1, GPT-6 Astra, Grok 4.6, and an open-source frontier model: Kimi-K3. The huge boost in Tables~\ref{tab:sota_likert}, \ref{tab:sota_videosim}, \ref{tab:sota_codegen} show that these frontier models have surpassed the models from the main paper. Kimi K3 does a really good job in coding and visually similar video generation, which is a win for open-source models, however with much more thinking token generation than the others. GPT-6-Astra is also good at visually similar video generation, however, Claude-Fable-5.1 seems to have the most physics understanding among the evaluated models.

\begin{table}[ht]
\centering
\small
\caption{\textbf{Average Likert scores} (on a 1--5 scale) rated by 3 LLM judge models (Gemini-2.5-flash,
Grok-4.1-Fast, GPT-5-Nano) for model outputs across physical law correctness and equivalence
dimensions, on the 10-sample hard subset.}
\label{tab:sota_likert}
\begin{tabular}{l|cc|cccc}
\hline
 & \multicolumn{2}{c|}{\textbf{Physical Law Correctness}} & \multicolumn{4}{c}{\textbf{Physical Law Equivalence}} \\
\textbf{Model} & Formula & Overall & Name & Statement & Formula & Overall \\
\hline
\multicolumn{7}{l}{\textit{New frontier models}} \\
\textsc{Claude Fable 5.1}          & 4.40 & 4.33 & 3.47 & \textbf{3.13} & \textbf{2.90} & \textbf{3.00} \\
\textsc{GPT-6 Astra}               & \textbf{4.73} & \textbf{4.50} & 3.27 & 2.87 & 2.63 & 2.80 \\
\textsc{Grok 4.6}                  & 4.40 & 4.23 & 2.87 & 2.53 & 2.30 & 2.40 \\
\textsc{Kimi K3}                   & 4.40 & 4.37 & \textbf{3.57} & 2.63 & 2.40 & 2.67 \\
\hline
\multicolumn{7}{l}{\textit{Open-source models}} \\
\textsc{Qwen3-VL-30B-A3B-Instruct} & 4.73 & 4.87 & 1.83 & 1.67 & 1.50 & 1.60 \\
\textsc{InternVL3.5-30B-A3B}       & 4.53 & 4.53 & 1.47 & 1.73 & 1.17 & 1.40 \\
\textsc{Gemma-3-12B-IT}            & \textbf{5.00} & \textbf{5.00} & \textbf{2.10} & 1.87 & 1.43 & 1.73 \\
\textsc{GLM-4.1V-9B-Thinking}      & 4.50 & 4.60 & 1.90 & \textbf{1.90} & 1.70 & \textbf{1.80} \\
\textsc{Pixtral-12B-2409}          & 4.63 & 4.63 & 1.80 & 1.50 & \textbf{1.83} & 1.53 \\
\hline
\multicolumn{7}{l}{\textit{Closed-source models}} \\
\textsc{GPT-5 Mini}                & \textbf{4.87} & 4.60 & 1.87 & 2.07 & 1.53 & 1.60 \\
\textsc{Gemini 2.5 Pro}            & 4.19 & 4.30 & 2.07 & 1.85 & \textbf{1.74} & \textbf{1.96} \\
\textsc{Claude Sonnet 4.6}         & 3.73 & 3.50 & \textbf{2.17} & \textbf{2.10} & 1.43 & 1.77 \\
\textsc{Grok 4 Fast Reasoning}     & 4.77 & 4.53 & 2.03 & \textbf{2.10} & 1.70 & 1.93 \\
\textsc{Nova 2 Lite}               & 4.80 & \textbf{4.63} & 2.10 & 1.93 & 1.70 & 1.80 \\
\hline
\end{tabular}
\end{table}

\begin{table}[t]
\centering
\small
\caption{\textbf{Video simulation equivalence table} containing verifiable video frame similarity scores
using DINOv2 and VideoCLIP, on the 10-sample hard subset. Cosine similarities on $[0,1]$, averaged over
successfully rendered videos (scipy and kubric pooled). Human-judgment columns are not available for this subset.}
\label{tab:sota_videosim}
\begin{tabular}{l|cc}
\hline
 & \multicolumn{2}{c}{\textbf{Patchwise Video Similarity}} \\
\textbf{Model} & DINOv2 sim. & VideoCLIP sim. \\
\hline
\multicolumn{3}{l}{\textit{New frontier models}} \\
\textsc{Claude Fable 5.1}          & 0.855 & 0.881 \\
\textsc{GPT-6 Astra}               & \textbf{0.902} & 0.896 \\
\textsc{Grok 4.6}                  & 0.781 & 0.869 \\
\textsc{Kimi K3}                   & 0.863 & \textbf{0.903} \\
\hline
\multicolumn{3}{l}{\textit{Open-source models}} \\
\textsc{Qwen3-VL-30B-A3B-Instruct} & 0.506 & 0.716 \\
\textsc{InternVL3.5-30B-A3B}       & 0.580 & 0.733 \\
\textsc{Gemma-3-12B-IT}            & \textbf{0.595} & 0.609 \\
\textsc{GLM-4.1V-9B-Thinking}      & 0.576 & \textbf{0.740} \\
\textsc{Pixtral-12B-2409}          & 0.339 & 0.669 \\
\hline
\multicolumn{3}{l}{\textit{Closed-source models}} \\
\textsc{GPT-5 Mini}                & 0.714 & 0.812 \\
\textsc{Gemini 2.5 Pro}            & 0.788 & \textbf{0.843} \\
\textsc{Claude Sonnet 4.6}         & \textbf{0.802} & 0.841 \\
\textsc{Grok 4 Fast Reasoning}     & 0.625 & 0.768 \\
\textsc{Nova 2 Lite}               & 0.428 & 0.666 \\
\hline
\end{tabular}
\end{table}

\begin{table}[t]
\centering
\small
\caption{\textbf{Evaluation of parameter estimation accuracy and generated simulation code quality} on the
10-sample hard subset (77 ground-truth parameters). Percentages computed as a fraction of all $N=10$ inputs
that successfully reached each stage (CoT extracted, Code extracted, Compiles, Runs without error, Saves video).
CodeBLEU averaged over scored samples.}
\label{tab:sota_codegen}
\resizebox{\linewidth}{!}{%
\begin{tabular}{l|c|cccccc}
\hline
 & Param. Estimation & \multicolumn{6}{c}{\textbf{Code Generation and Execution}} \\
\textbf{Model} & Acc. ($\pm$20\%) & CoT & Code & Compile & No Errors & Video Save & CodeBLEU \\
\hline
\multicolumn{8}{l}{\textit{New frontier models}} \\
\textsc{Claude Fable 5.1}          & \textbf{18.18\%} & \textbf{100.0} & \textbf{100.0} & \textbf{100.0} & \textbf{100.0} & \textbf{100.0} & 0.240 \\
\textsc{GPT-6 Astra}               & 5.19\%  & \textbf{100.0} & \textbf{100.0} & \textbf{100.0} & \textbf{100.0} & \textbf{100.0} & 0.240 \\
\textsc{Grok 4.6}                  & 9.09\%  & \textbf{100.0} & \textbf{100.0} & \textbf{100.0} & \textbf{100.0} & \textbf{100.0} & 0.237 \\
\textsc{Kimi K3}                   & 12.99\% & \textbf{100.0} & \textbf{100.0} & \textbf{100.0} & \textbf{100.0} & \textbf{100.0} & \textbf{0.244} \\
\hline
\multicolumn{8}{l}{\textit{Open-source models}} \\
\textsc{Qwen3-VL-30B-A3B-Instruct} & \textbf{3.90\%} & \textbf{100.0} & \textbf{100.0} & \textbf{100.0} & \textbf{80.0} & \textbf{80.0} & \textbf{0.202} \\
\textsc{InternVL3.5-30B-A3B}       & \textbf{3.90\%} & \textbf{100.0} & 90.0 & 60.0 & 10.0 & 10.0 & 0.184 \\
\textsc{Gemma-3-12B-IT}            & \textbf{3.90\%} & \textbf{100.0} & \textbf{100.0} & \textbf{100.0} & 10.0 & 10.0 & 0.193 \\
\textsc{GLM-4.1V-9B-Thinking}      & \textbf{3.90\%} & \textbf{100.0} & 70.0 & 70.0 & 30.0 & 30.0 & 0.189 \\
\textsc{Pixtral-12B-2409}          & \textbf{3.90\%} & \textbf{100.0} & \textbf{100.0} & \textbf{100.0} & 50.0 & 60.0 & 0.179 \\
\hline
\multicolumn{8}{l}{\textit{Closed-source models}} \\
\textsc{GPT-5 Mini}                & 3.90\% & 50.0 & 90.0 & 90.0 & 80.0 & 80.0 & \textbf{0.245} \\
\textsc{Gemini 2.5 Pro}            & 5.19\% & 90.0 & 60.0 & 60.0 & 60.0 & 60.0 & 0.199 \\
\textsc{Claude Sonnet 4.6}         & 5.19\% & \textbf{100.0} & 90.0 & 90.0 & \textbf{90.0} & \textbf{90.0} & 0.232 \\
\textsc{Grok 4 Fast Reasoning}     & \textbf{6.49\%} & \textbf{100.0} & \textbf{100.0} & \textbf{100.0} & \textbf{90.0} & \textbf{90.0} & 0.196 \\
\textsc{Nova 2 Lite}               & 2.60\% & \textbf{100.0} & \textbf{100.0} & \textbf{100.0} & 70.0 & 80.0 & 0.195 \\
\hline
\end{tabular}%
}
\end{table}

\subsection{Rigorous evaluation to measure Claude bias}
\label{app:claude_bias}

\begin{table*}[ht]
\centering
\small
\caption{Parameter-estimation evaluation across 10 MLLMs on the 2{,}430 benchmark samples (162 experiments $\times$ 15 samples). \textbf{Parse fails}: number of samples in which the model's response could not be parsed into a valid CoT block (no parameters recovered for those samples). \textbf{Avg pred params}: mean number of physical
parameters the model emits per sample. \textbf{Avg matched}: mean number of model parameters that were successfully aligned to a ground-truth parameter, by greedy bipartite matching over alias sets that capture deabbreviation and Greek/Latin synonyms. \textbf{Mean frac reported}: mean over rows of $n_{\text{matched}}/n_{\text{gt}}$,
i.e.\ recall on \emph{naming} the correct parameters. \textbf{Mean frac within $\mathbf{20\%}$}: mean over rows of $n_{\text{within tol}}/n_{\text{matched}}$, i.e.\ \emph{value}-accuracy among matched parameters at $\pm 20\%$ relative-error tolerance.
\textbf{Global accuracy}: $\sum n_{\text{within tol}} / \sum n_{\text{matched}}$ pooled across all 2{,}430 rows. \textbf{End-to-end recovery}: $\sum n_{\text{within tol}} / \sum n_{\text{gt}}$, the fraction of \emph{all} ground-truth parameters that the model both named correctly and estimated within $\pm 20\%$. Best per column in \textbf{bold}.}
\label{tab:param-eval-detailed}
\setlength{\tabcolsep}{4pt}
\begin{tabular}{lrrrrrrr}
\toprule
Model & Parse & Avg pred & Avg & Mean frac & Mean frac & Global & End-to-end \\
      & fails & params   & matched & reported & within $20\%$ & accuracy & recovery \\
\midrule
\texttt{claude\_sonnet\_4\_6}      & \textbf{0}   & 7.7        & \textbf{3.29} & \textbf{0.460} & 0.171 & 0.185          & \textbf{0.0769} \\
\texttt{gemini\_2\_5\_pro}         & 96           & 5.5        & 2.83          & 0.399          & 0.150 & 0.170          & 0.0606          \\
\texttt{gemma\_3\_12b\_it}         & 330          & 4.8        & 1.94          & 0.281          & 0.130 & 0.163          & 0.0398          \\
\texttt{glm\_4\_1v\_9b\_thinking}  & 420          & 4.0        & 1.93          & 0.273          & 0.147 & \textbf{0.199} & 0.0484          \\
\texttt{gpt\_5\_mini}              & 465          & \textbf{8.4} & 3.17        & 0.426          & 0.139 & 0.178          & 0.0713          \\
\texttt{grok\_4\_fast\_reasoning}  & 32           & 5.9        & 2.74          & 0.389          & 0.155 & 0.165          & 0.0569          \\
\texttt{internvl3\_5\_30b\_a3b}    & 22           & 3.7        & 1.77          & 0.259          & 0.141 & 0.158          & 0.0352          \\
\texttt{nova\_2\_lite}             & 203          & 4.0        & 1.78          & 0.258          & 0.136 & 0.167          & 0.0376          \\
\texttt{pixtral\_12b}              & 365          & 3.5        & 1.39          & 0.198          & 0.134 & 0.188          & 0.0331          \\
\texttt{qwen3vl\_30b\_a3b}         & 29           & 6.2        & 2.80          & 0.407          & \textbf{0.158} & 0.174 & 0.0616          \\
\bottomrule
\end{tabular}
\end{table*}

We request to observe a few facts: (i) Claude Sonnet 4.6 leads 0 of the 11 physics/similarity metrics in Tables \ref{tab:likert_scores} and \ref{tab:video-sim-equivalence}. (ii) No Claude model was on the judge panel deliberately, so self-bias was not there during evaluation. (iii) In Table \ref{tab:code_generation_metrics}, Claude leads in some code generation and execution reliability metrics, but it can’t be due to a GT-authorship bias; it’s just Claude being a strong coding model. And Claude leads in the parameter/ estimation accuracy; actually, it comes from naming recall (0.460, best of 10) rather than per-parameter value accuracy, where it ranks only 3rd (0.185, behind GLM-4.1V 0.199 and Pixtral 0.188) (Table \ref{tab:param-eval-detailed}), so breadth of coverage, not better physics, and not authorship.

To validate the debiasing, we rebuilt a matched subset with GPT-5.5: 15 SciPy experiments × 5 samples = 75 samples. Everything is GPT-generated this time, including parameter choices and sampling ranges, through the identical curation pipeline with the same 2-10 rounds of human physical-plausibility iteration. This GPT arm is not debiased (only mandatory preprocessing: comment stripping, CoT normalisation to LaTeX/Python), while the Claude arm was already debiased. We then ran 8 MLLMs (3 closed + 5 open) on both arms.

\begin{table}[t]
\centering
\small
\caption{Bias scores across evaluation metrics for Claude Sonnet 4.6 and GPT-5 Mini. The gap is computed as GPT-5 Mini minus Claude Sonnet 4.6.}
\label{tab:bias_scores}
\begin{tabular}{lccc}
\toprule
\textbf{Metric} & \textbf{Claude Sonnet 4.6} & \textbf{GPT-5 Mini} & \textbf{Gap (GPT $-$ Claude)} \\
\midrule
Law equivalence -- Overall      & +0.27    & +0.30    & +0.03    \\
Law validity -- Overall         & +0.02    & $-0.04$  & $-0.06$  \\
\textbf{CodeBLEU}               & \textbf{+0.000} & $-0.003$ & $-0.003$ \\
Runs without error              & $-3.2$ pp & $-3.9$ pp & $-0.7$ pp \\
Video save                      & $-2.2$ pp & $-1.6$ pp & +0.6 pp \\
\textbf{Param. Est. Acc. ($\pm20\%$)}
                                & \textbf{+0.76 pp} & \textbf{$-1.93$ pp} & \textbf{$-2.69$ pp} \\
\midrule
\textit{If the benchmark favored Claude}
                                & \textit{large negative} & \textit{large positive} & \textit{large positive} \\
\bottomrule
\end{tabular}
\end{table}

For each of Claude and GPT (as inference models), we measure how much its score moves when the ground truth switches from Claude-authored to GPT-authored, then subtract the average move of the 6 neutral models (neither Claude- nor GPT-family). Subtracting the neutral models cancels out the fact that the two arms simply differ in difficulty, so what remains is the family-specific effect only. We call this the bias score (formally, difference-in-differences). A positive bias score means the model does relatively better on GPT-authored ground truth; a negative score means it does better on Claude-authored ground truth. So if Claude bias existed, Claude's bias score would be clearly negative (it loses its home advantage) and GPT-5 Mini's clearly positive. Neither happens (Table \ref{tab:bias_scores}).

Claude's bias score on parameter estimation is positive (+0.76 pp); it does better on GPT-authored ground truth. GPT-5 Mini's is negative (-1.93 pp); it does worse on its own family's ground truth. On CodeBLEU, the one metric that literally string-compares model output against the ground-truth code and is therefore most exposed to authorship style, Claude's bias score is exactly 0.000, and GPT-5 Mini beats Claude on both arms (0.203 / 0.219 vs 0.189 / 0.201). On the Claude-authored arm, GPT-5 Mini wins parameter recovery (7.81\%) and Claude places only 3rd (6.30\%); on the GPT-authored arm, Claude leads (7.52\%). A benchmark that favored its creator model could not place that model 3rd on its own ground truth.

Finally, the two arms rank the 8 models almost identically. In Table \ref{ranking_sim}, $\rho = +1$ means the two arms produce the same ranking of the 8 models, and $\rho$ = 0 means the rankings are unrelated. $\rho$ is positive on all 13 metrics (mean +0.775). $p$ is the standard significance value for each correlation: the probability of getting a $\rho$ that large purely by luck, if the two arms actually ranked the 8 models in unrelated orders. Small $p$ = the agreement is real; large $p$ = it could easily be coincidence. We restricted the counterfactual arm to SciPy because SciPy GT code is free-form, so generator style has maximum room to leave a signature, whereas Kubric code is written against a fixed 5-block skeleton, which suppresses authorship style; a Kubric arm would bound the bias lower, not higher. Table \ref{tab:param-eval-gpt-detailed} shows the parameter estimation evaluation comparison; Table~\ref{tab:gpt_likert_scores} shows the physical law correctness and equivalence scores (scored by the same LLM judge panel); and Table~\ref{tab:gpt_code_generation_metrics} shows the $\pm20\%$ parameter estimation accuracy along with code evaluation metric values of 5 open-source and 3 closed-source benchmarking models on the smaller 75-sample subset authored by Claude and GPT. 

\begin{table*}[t]
\centering
\footnotesize
\setlength{\tabcolsep}{0.4em}
\caption{\textbf{Average Likert scores} (on a 1--5 scale) rated by 3 LLM judge models (Gemini-2.5-flash, Grok-4.1-Fast, GPT-5-Nano) for model outputs on a matched subset of 75 samples per arm — 15 Scipy experiments with 5 samples each, generated by \textbf{GPT-5.5 (left) vs. Claude Opus 4.7 (right)} within each cell as ``GPT arm / Claude arm''.}
\label{tab:gpt_likert_scores}
\begin{adjustbox}{width=\linewidth}
\begin{tabular}{l|cc|cccc}
\toprule
& \multicolumn{2}{c|}{\textbf{Physical Law Correctness}} & \multicolumn{4}{c}{\textbf{Physical Law Equivalence}} \\
\textbf{Model} & \textbf{Formula} & \textbf{Overall} & \textbf{Name} & \textbf{Statement} & \textbf{Formula} & \textbf{Overall} \\
\midrule
\rowcolor{gray!15} \multicolumn{7}{l}{\textit{Open-source models}} \\
\textsc{Qwen3-VL-30B-A3B-Instruct} & 4.76 / 4.71 & 4.75 / 4.77 & 2.08 / 2.00 & 2.10 / 1.95 & 1.58 / 1.56 & 1.81 / 1.80 \\
\textsc{InternVL3.5-30B-A3B}       & 4.67 / 4.67 & 4.45 / 4.56 & 1.75 / 1.83 & 1.84 / 2.04 & 1.35 / 1.53 & 1.53 / 1.73 \\
\textsc{Gemma-3-12B-IT}            & 4.57 / 4.74 & 4.69 / 4.81 & 1.77 / 2.49 & 1.70 / 2.22 & 1.29 / 1.59 & 1.48 / 1.99 \\
\textsc{GLM-4.1V-9B-Thinking}      & 4.29 / 4.44 & 4.08 / 4.32 & 1.99 / 1.98 & 2.00 / 1.97 & 1.45 / 1.54 & 1.69 / 1.74 \\
\textsc{Pixtral-12B-2409}          & 4.57 / 4.61 & 4.46 / 4.59 & 1.88 / 2.12 & 1.61 / 1.85 & 1.32 / 1.50 & 1.49 / 1.73 \\
\midrule
\rowcolor{gray!15} \multicolumn{7}{l}{\textit{Closed-source models}} \\
\textsc{GPT-5 Mini}                & 4.67 / 4.72 & 4.50 / 4.68 & 2.69 / 2.46 & 2.65 / 2.46 & 2.19 / 1.96 & 2.35 / 2.19 \\
\textsc{Gemini 2.5 Pro}            & 4.51 / 4.71 & 4.34 / 4.59 & 2.13 / 2.10 & 2.28 / 2.09 & 1.90 / 1.75 & 2.02 / 1.89 \\
\textsc{Claude Sonnet 4.6}         & 4.43 / 4.56 & 4.28 / 4.40 & 2.55 / 2.43 & 2.52 / 2.38 & 1.95 / 1.86 & 2.22 / 2.09 \\
\bottomrule
\end{tabular}
\end{adjustbox}
\end{table*}

% --- Code Generation ---
\begin{table*}[t]
\centering
\footnotesize
\setlength{\tabcolsep}{0.4em}
\caption{\textbf{Evaluation of parameter estimation accuracy and generated simulation code quality on the matched 75-sample subset}. Each cell reports \textbf{GPT arm / Claude arm} on 15 Scipy experiments with 5 samples each. Percentages computed as a fraction of $N=75$ inputs per arm that successfully reached each stage. CodeBLEU averaged over scored samples. Param.\ Estimation Acc.\ uses alias matching (deabbreviation + Greek/Latin synonyms) with $\pm20\%$ tolerance.}
\label{tab:gpt_code_generation_metrics}
\begin{adjustbox}{width=\linewidth}
\begin{tabular}{l|c|cccccc}
\toprule
& \textbf{Param. Estimation} & \multicolumn{6}{c}{\textbf{Code Generation and Execution}} \\
\textbf{Model} & \textbf{Acc. (\(\pm20\%\))} & \textbf{CoT} & \textbf{Code} & \textbf{Compile} & \textbf{No Errors} & \textbf{Video Save} & \textbf{CodeBLEU} \\
\midrule
\rowcolor{gray!15} \multicolumn{8}{l}{\textit{Open-source models}} \\
\textsc{Qwen3-VL-30B-A3B-Instruct} & 7.33\% / 4.52\% & 100.0\% / 100.0\% & 96.0\% / 100.0\% & 96.0\% / 98.7\% & 60.0\% / 58.7\% & 70.7\% / 64.0\% & 0.142 / 0.156 \\
\textsc{InternVL3.5-30B-A3B}       & 4.36\% / 4.25\% & 100.0\% / 100.0\% & 98.7\% / 100.0\% & 98.7\% / 100.0\% & 42.7\% / 42.7\% & 60.0\% / 65.3\% & 0.125 / 0.137 \\
\textsc{Gemma-3-12B-IT}            & 2.57\% / 2.88\% & 98.7\% / 100.0\% & 98.7\% / 94.7\% & 98.7\% / 94.7\% & 22.7\% / 20.0\% & 33.3\% / 34.7\% & 0.137 / 0.145 \\
\textsc{GLM-4.1V-9B-Thinking}      & 6.73\% / 6.99\% & 100.0\% / 97.3\% & 100.0\% / 97.3\% & 100.0\% / 96.0\% & 57.3\% / 48.0\% & 58.7\% / 50.7\% & 0.142 / 0.152 \\
\textsc{Pixtral-12B-2409}          & 3.17\% / 4.25\% & 100.0\% / 100.0\% & 100.0\% / 100.0\% & 100.0\% / 100.0\% & 40.0\% / 33.3\% & 73.3\% / 72.0\% & 0.118 / 0.131 \\
\midrule
\rowcolor{gray!15} \multicolumn{8}{l}{\textit{Closed-source models}} \\
\textsc{GPT-5 Mini}                & 6.34\% / 7.81\% & 90.7\% / 89.3\% & 97.3\% / 93.3\% & 97.3\% / 93.3\% & 84.0\% / 80.0\% & 84.0\% / 80.0\% & 0.203 / 0.219 \\
\textsc{Gemini 2.5 Pro}            & 5.74\% / 4.25\% & 96.0\% / 97.3\% & 73.3\% / 81.3\% & 73.3\% / 81.3\% & 70.7\% / 73.3\% & 70.7\% / 76.0\% & 0.165 / 0.181 \\
\textsc{Claude Sonnet 4.6}         & 7.52\% / 6.30\% & 100.0\% / 100.0\% & 98.7\% / 98.7\% & 98.7\% / 98.7\% & 94.7\% / 93.3\% & 94.7\% / 94.7\% & 0.189 / 0.201 \\
\bottomrule
\end{tabular}
\end{adjustbox}
\end{table*}

\begin{table}[t]
\centering
\small
\caption{Model ranking similarity score measured by Spearman's rank correlation coefficient ($\rho$) and corresponding significance value ($p$) for all the evaluation metrics}
\label{ranking_sim}
\begin{tabular}{lcc}
\toprule
\textbf{Metric} & $\boldsymbol{\rho}$ & $\mathbf{p}$ \\
\midrule
CodeBLEU                    & +1.000          & $\sim$0 \\
Law validity - Statement    & +0.976          & $\sim$0 \\
Runs without error          & +0.976          & $\sim$0 \\
Law validity - Name         & +0.952          & $\sim$0 \\
Law validity - Overall      & +0.905          & 0.002 \\
Video save                  & +0.905          & 0.002 \\
Param. Est. Acc. ($\pm20\%$) & +0.786        & 0.021 \\
Law equivalence - Formula   & +0.762          & 0.028 \\
Law equivalence - Statement & +0.667          & 0.071 \\
Law equivalence - Overall   & +0.643          & 0.086 \\
Syntax OK                   & +0.643          & 0.086 \\
Law validity - Formula      & +0.500          & 0.207 \\
Law equivalence - Name      & +0.357          & 0.385 \\
\midrule
\textbf{All 13 metrics}     & \textbf{Mean +0.775} & \textbf{13/13 positive} \\
\bottomrule
\end{tabular}
\end{table}

\begin{table*}[ht]
\centering
\footnotesize
\caption{\textbf{Parameter-estimation evaluation split by GT authorship}, on the matched 75-sample subset (15 Scipy experiments $\times$ 5 samples). Each cell reports \textbf{GPT arm / Claude arm} using the same alias-matcher as Table \ref{tab:param-eval-detailed}. Column definitions unchanged.}
\label{tab:param-eval-gpt-detailed}
\setlength{\tabcolsep}{3pt}
\begin{adjustbox}{width=\linewidth}
\begin{tabular}{lrrrrrrr}
\toprule
Model & Parse & Avg pred & Avg & Mean frac & Mean frac & Global & End-to-end \\
      & fails & params   & matched & reported & within $20\%$ & accuracy & recovery \\
\midrule
\texttt{claude\_sonnet\_4\_6}      & 0 / 0    & 7.7 / 7.3   & 3.07 / 3.45 & 0.459 / 0.346 & 0.176 / 0.164 & 0.165 / 0.178 & 0.0752 / 0.0630 \\
\texttt{gemini\_2\_5\_pro}         & 3 / 2    & 5.9 / 5.3   & 2.51 / 2.68 & 0.378 / 0.284 & 0.137 / 0.175 & 0.154 / 0.154 & 0.0574 / 0.0425 \\
\texttt{gemma\_3\_12b\_it}         & 21 / 34  & 4.6 / 3.6   & 1.75 / 1.67 & 0.268 / 0.181 & 0.063 / 0.114 & 0.099 / 0.168 & 0.0257 / 0.0288 \\
\texttt{glm\_4\_1v\_9b\_thinking}  & 6 / 11   & 5.2 / 4.7   & 2.59 / 2.59 & 0.385 / 0.272 & 0.141 / 0.228 & 0.175 / 0.263 & 0.0673 / 0.0699 \\
\texttt{gpt\_5\_mini}              & 7 / 10   & 10.5 / 9.4  & 3.63 / 3.87 & 0.544 / 0.422 & 0.116 / 0.168 & 0.118 / 0.197 & 0.0634 / 0.0781 \\
\texttt{internvl3\_5\_30b\_a3b}    & 0 / 0    & 4.0 / 3.9   & 1.57 / 1.81 & 0.235 / 0.195 & 0.131 / 0.223 & 0.186 / 0.228 & 0.0436 / 0.0425 \\
\texttt{pixtral\_12b}              & 9 / 0    & 4.0 / 4.4   & 1.25 / 1.99 & 0.183 / 0.204 & 0.104 / 0.206 & 0.170 / 0.208 & 0.0317 / 0.0425 \\
\texttt{qwen3vl\_30b\_a3b}         & 0 / 1    & 6.1 / 5.8   & 2.61 / 2.76 & 0.413 / 0.297 & 0.156 / 0.145 & 0.189 / 0.159 & 0.0733 / 0.0452 \\
\bottomrule
\end{tabular}
\end{adjustbox}
\end{table*}

\subsection{Statistical Justification of choosing Subset for Benchmarking}
\label{app:stat}
Within an experiment, samples are i.i.d.\ in parameter values with the physics held constant, so the per-experiment success rate is a binomial proportion whose standard error at $n=15$ is bounded by $\sqrt{0.25/15}\approx 0.129$ assuming the maximum possible variance for a Bernoulli. Averaging across 162 independent experiments bounds the per-model aggregate standard error to $0.129/\sqrt{162}\approx 0.010$, i.e.\ a 95\% confidence interval of $\pm 2\%$ on every metric reported in Tables \ref{tab:likert_scores}--\ref{tab:code_generation_metrics}. This 95\% CI bound on every reported metric (±0.010) statistically supports every between-model gap we report (e.g., $7.7\%$ vs.\ $3.3\%$ on parameter recovery), so the relative rankings established on the 2,430-sample subset are statistically supported and very likely to hold on the full 160,614-sample corpus, while API cost remains tractable.

\subsection{Inference Prompt}
\label{app:prompt}

\begin{table}[b]
\centering
\footnotesize
\setlength{\tabcolsep}{0.4em}
\caption{\textbf{Chain-of-thought schema.} All samples contain simulation, physical\_observation, physical\_law, ode\_system, derivation\_steps, numerical\_method, code\_validation, expected\_behavior and parameters; the remaining keys appear only where applicable, and SciPy samples store physical constants as physical\_constants\_derived.}
\label{tab:cot_schema}
\begin{adjustbox}{width=\linewidth}
\begin{tabular}{ll}
    \toprule
    \textbf{Key} & \textbf{Content} \\
    \midrule
    \texttt{simulation} & Experiment-type identifier. \\
    \texttt{category} & Physics domain (e.g., articulated, rigid-body collision). \\
    \texttt{physical\_observation} & Natural-language description of the visual dynamics, with parameter values. \\
    \texttt{physical\_law} & Law name, statement, and governing equation in \LaTeX. \\
    \texttt{modeling\_assumptions} & Idealisations baked in (point masses, frictionless surfaces, etc.). \\
    \texttt{ode\_system} & State variables and governing ODE in both \LaTeX{} and Python form. \\
    \texttt{event\_schedule} & Discrete events for collisions / contacts (only when applicable). \\
    \texttt{derivation\_steps} & Ordered derivation from first principles to the implemented update. \\
    \texttt{physical\_constants} & Computed quantities (moments of inertia, equilibrium positions, etc.). \\
    \texttt{numerical\_method} & Solver, tolerances, and closed-form availability. \\
    \texttt{code\_validation} & Analytical checks, limiting behaviours, and plausibility flags. \\
    \texttt{expected\_behavior} & Qualitative predictions about the resulting trajectory. \\
    \texttt{parameters} & All sampled values with units and human-readable descriptions. \\
    \bottomrule
\end{tabular}
\end{adjustbox}
\end{table}

Every model under evaluation receives the engine-conditional prompt template
shown in Listing~\ref{lst:prompt-template}. The placeholders
\texttt{\{engine\_label\}}, \texttt{\{schema\}}, and \texttt{\{code\_rules\}}
are filled with the values of Listings~\ref{lst:scipy-schema}--\ref{lst:kubric-rules}
when the engine is \texttt{scipy} or \texttt{kubric}, respectively. The two
delimited output blocks (\texttt{@@COT\_BEGIN@@\ldots@@COT\_END@@} and
\texttt{@@CODE\_BEGIN@@\ldots@@CODE\_END@@}) are parsed deterministically;
parse failures count toward the ``parse fails'' column of
Table~\ref{tab:param-eval-detailed}.

% ------------------------------------------------------------------
\begin{lstlisting}[
  style   = prompt,
  caption = {Engine-conditional prompt template (Python f-string source).
             The same template is sent to every model; \texttt{\{engine\_label\}},
             \texttt{\{schema\}}, and \texttt{\{code\_rules\}} are interpolated
             from Listings~\ref{lst:scipy-schema}--\ref{lst:kubric-rules}.},
  label   = {lst:prompt-template}
]
You are a physics simulation expert. You will be shown a video
(10 FPS, duration-preserving) of a physical scene. The simulation
engine is: {engine_label}.

Your task: from the video alone, infer the physics and output BOTH:
  (1) a Chain-of-Thought analysis as a JSON object (schema below), and
  (2) a self-contained pure-Python script (see CODE REQUIREMENTS for the
      exact libraries) that, when executed, reproduces the dynamics
      and writes the rendered video to video.mp4.

== OUTPUT PROTOCOL (strict) ==
Emit exactly two blocks, each delimited by the markers below. Do NOT
wrap them in code fences. Do NOT add any text outside the markers.

@@COT_BEGIN@@
<one valid JSON object matching the schema below>
@@COT_END@@

@@CODE_BEGIN@@
<full Python source code, raw (no triple backticks)>
@@CODE_END@@

== CHAIN-OF-THOUGHT JSON SCHEMA ==
The JSON object MUST use these exact top-level keys (and nested keys
where shown). Use null where not applicable.
{schema}

== CODE REQUIREMENTS ==
{code_rules}

Begin now. Output only the two delimited blocks, nothing else.
\end{lstlisting}

% ------------------------------------------------------------------
\begin{lstlisting}[
  style   = prompt,
  caption = {Chain-of-thought JSON schema for the \texttt{scipy} engine.},
  label   = {lst:scipy-schema}
]
{
  simulation: name, eg: falling objects,
  physical_observation: <one paragraph: what you see, geometry, key numerical scales>,
  physical_law: {
    name: <law / theorem name>,
    statement: <one-sentence statement>,
    formula: <LaTeX formula>
  },
  parameters: {
    <name>: {value: <number>, unit: <SI unit or empty>, description: <short>},
    <...>: <...>
  }
}
\end{lstlisting}

% ------------------------------------------------------------------
\begin{lstlisting}[
  style   = prompt,
  caption = {Chain-of-thought JSON schema for the \texttt{kubric} engine.
             Identical structure to the \texttt{scipy} schema; only the
             \texttt{simulation} description changes to reflect 3-D rigid-body
             scenes.},
  label   = {lst:kubric-schema}
]
{
  simulation: name, eg: falling object,
  physical_observation: <one paragraph: objects, geometry, key parameter values>,
  physical_law: {
    name: <law name>,
    statement: <one sentence>,
    formula: <LaTeX>
  },
  parameters: {
    <name>: {value: <number>, unit: <units>, description: <short>},
    <...>: <...>
  }
}
\end{lstlisting}

% ------------------------------------------------------------------
\begin{lstlisting}[
  style   = prompt,
  caption = {Code-generation rules for the \texttt{scipy} engine. The model
             must emit a self-contained Python script that uses only stdlib,
             NumPy, SciPy, and Matplotlib's \texttt{FuncAnimation} to render
             the predicted dynamics into \texttt{video.mp4}.},
  label   = {lst:scipy-rules}
]
- Write PURE PYTHON that runs in a standard Python environment
  (no Docker, no external physics engines).
- Use ONLY stdlib + numpy + scipy + matplotlib. Do NOT import any 3D engine.
- Use this exact import block (you may omit a line you don't need;
  do not add other heavy deps):
    import numpy as np
    import matplotlib.pyplot as plt
    import matplotlib.animation as animation
    from matplotlib.patches import Polygon, Rectangle, Circle
    from matplotlib.collections import LineCollection
    from scipy.integrate import solve_ivp
- Solve dynamics analytically when a closed form exists (kinematics,
  damped oscillators); otherwise call scipy.integrate.solve_ivp with
  RK45 and tight tolerances (rtol=1e-9, atol=1e-11).
- Render with matplotlib.animation.FuncAnimation: build patches/lines
  once, then update them via set_xy / set_data / set_segments inside
  the update(frame) function. Use blit=False if state is complex.
- IMPORTANT: assign FuncAnimation to a top-level variable (e.g.
  ani = animation.FuncAnimation(...)) and then call ani.save(...).
  Do NOT chain inline like animation.FuncAnimation(...).save(...) -
  matplotlib garbage-collects the unreferenced animation and silently
  writes an empty / corrupt video.
- Save with: ani.save('video.mp4',
      writer=animation.FFMpegWriter(fps=30, bitrate=1800), dpi=120).
  Output file MUST be exactly 'video.mp4' in the current working
  directory.
- Render at fps=30. Match the duration of the observed video as best
  you can infer.
- The script must be standalone (no command-line args, no relative
  file deps) and runnable as `python simulation_code.py`.
- DO NOT wrap the code in ```python ... ``` triple-backtick fences.
  Emit raw Python ONLY between the @@CODE_BEGIN@@ and @@CODE_END@@
  markers, no fences anywhere.
\end{lstlisting}

% ------------------------------------------------------------------
\begin{lstlisting}[
  style   = prompt,
  caption = {Code-generation rules for the \texttt{kubric} engine. The
             model must emit a self-contained PyBullet + pyrender + trimesh
             + imageio Python script (the literal \texttt{kubric} package is
             out of scope because it requires Blender + Docker). The rules
             list hard environment bans, an exact 5-block skeleton, the
             keyword signatures of every PyBullet API the script may call,
             and a self-check the model is asked to perform before emitting.
             For brevity we reproduce the abridged version.},
  label   = {lst:kubric-rules}
]
Write PURE PYTHON. Do NOT use Docker. Do NOT `import kubric` - that
package requires Blender + Docker and is OUT OF SCOPE. 'kubric' here
is the 3D-rigid-body category; implement with PyBullet (physics) +
pyrender (offscreen rendering) + trimesh + imageio.

== HARD ENVIRONMENT BANS ==
- trimesh.creation.sphere(...) -> use trimesh.creation.icosphere(
  subdivisions=3, radius=R). Allowed shapes: box, icosphere,
  cylinder, capsule. Anything else is forbidden.
- trimesh.visual.<anything> -> all forbidden. Color is set
  EXCLUSIVELY via pyrender.MetallicRoughnessMaterial(...).
- loadURDF(<anything other than 'plane.urdf'>) is forbidden. All
  shapes built with p.createCollisionShape(GEOM_BOX|SPHERE|CYLINDER|
  CAPSULE) + p.createMultiBody(...).
- p.createConstraint(...) is forbidden in this build.
- p.GEOM_MESH is forbidden.

== EXACT 5-BLOCK SKELETON ==
1. IMPORTS (do not add heavy deps beyond these):
   import os, math, random
   import numpy as np
   import pybullet as p
   import pybullet_data
   import trimesh, pyrender, imageio
2. PHYSICS SETUP (setAdditionalSearchPath BEFORE any loadURDF).
3. RENDERER + SCENE: at least one DirectionalLight AND exactly one
   PerspectiveCamera, both added before the frame loop.
4. WORLD: pybullet bodies + matching pyrender nodes (kept paired).
5. SIM LOOP -> SAVE: append to frames each iteration; write
   'video.mp4' at fps=30, 480x480.

== EXACT API SIGNATURES (use EXACT keyword names) ==
- p.createCollisionShape(GEOM_BOX, halfExtents=[hx,hy,hz],
                         physicsClientId=c)
- p.createCollisionShape(GEOM_SPHERE, radius=R, physicsClientId=c)
- p.createCollisionShape(GEOM_CYLINDER, radius=R, height=H,
                         physicsClientId=c)         # NOT length=
- p.createMultiBody(baseMass, baseCollisionShapeIndex,
                    baseVisualShapeIndex=-1,
                    basePosition=[x,y,z],
                    baseOrientation=[qx,qy,qz,qw],
                    physicsClientId=c)
- p.changeDynamics(body_id, -1, lateralFriction=...,
                   restitution=..., physicsClientId=c)
- p.resetBaseVelocity(body_id, linearVelocity=[...],
                      angularVelocity=[...], physicsClientId=c)
- p.applyExternalForce(body_id, -1, forceObj=[...],
                       posObj=[0,0,0], flags=p.LINK_FRAME,
                       physicsClientId=c)

== SELF-CHECK BEFORE EMITTING ==
(a) setAdditionalSearchPath precedes every loadURDF;
(b) no createJoint / createConstraint call exists;
(c) every PyBullet call passes physicsClientId=c;
(d) the frame loop appends to frames exactly once per iteration;
(e) the script ends with p.disconnect(c); renderer.delete().

Output file MUST be exactly `video.mp4` in the current working
directory; resolution 480x480, fps=30; runnable as
`python simulation_code.py` with no args.
\end{lstlisting}

\subsection{Human Rating Guidelines}
\label{app:guideline}
Raters were given coarse band definitions and asked to apply holistic judgment: for physical plausibility, 1-2 = physically impossible, 3 = broadly plausible with visible defects, 4 = minor defects, 5 = fully plausible; for equivalence (physics and appearance), 1 = unrelated, 2 = mostly different with a few matching elements, 3 = about half matching, 4 = mostly matching, 5 = closely matching. We deliberately kept the bands coarse because the failure modes vary widely across 162 phenomena. E.g. one regenerated video reproduced a correctly-colored ball bouncing at the right rhythm but with no floor present, which we scored 4 on appearance similarity and 2 on physical plausibility.

\subsection{Unobservable parameters and sensitivity stratification}
Different physical parameters have different extent of sensitivity. Sometimes, perturbing a parameter doesn't result into significant dynamics change, sometimes the video doesn't have enough information to estimate certain parameters (like experiments which have mass-independent dynamics). We ran a perturbation pilot study to categorize the sensitivity of the parameters.

For 15 experiments covering 108 physical parameters, we shifted each parameter by $\pm 10\%$, re-rendered the video, and measured how much the video changed (mean per-frame pixel MSE against the unperturbed video; $N = 1,080$ measurements). We heuristically define a parameter's sensitivity as that video change divided by the average video change of all parameters in the same experiment. So sensitivity = 1 means the parameter moves the video as much as a typical parameter in its own scene, 2 means twice as much, and 0.5 means half. Normalizing within an experiment is essential, because raw pixel change is dominated by how visually busy a scene is rather than by physics: it ranges 142$\times$ across experiments (0.000144 for \texttt{ball\_wrapping\_pole}, a single ball against a dark background, up to 0.0205 for \texttt{slab\_springs\_block\_drop}, which has many moving parts). After normalization that 142$\times$ spread collapses to 4.4$\times$.

\begin{table}[t]
\centering
\caption{Sensitivity across the 108 parameters}
\label{sensitivity}
\begin{tabular}{lc}
\toprule
\textbf{Statistical parameters} & \textbf{Value} \\
\midrule
Median                           & 0.98 \\
Middle 50\% (IQR)                & \textbf{[0.93, 1.05]} \\
90th / 95th percentile           & 1.17 / 1.22 \\
Full range                       & [0.48, 2.12] \\
Parameters above 1.5 / below 0.6 & 3 / 4 (of 108) \\
\bottomrule
\end{tabular}
\end{table}

The result in Table \ref{sensitivity} is that sensitivity is far more uniform than expected: half of all parameters sit within $\pm 5\%$ of their scene's average, and only 7 of 108 fall outside [0.6, 1.5]. In other words, at a fixed relative perturbation, most physical parameters in most scenes produce comparably-sized visual consequences. This speaks directly to why we chose a single fixed $\pm 20\%$ tolerance: because our threshold is also relative, and sensitivity at fixed relative perturbation is near-uniform, a single tolerance is much less crude than it first appears. We also found no parameter that is unobservable in the strict sense. The minimum normalized sensitivity is 0.48, roughly half the scene average, not zero; there is a sensitivity spectrum, but no degenerate cases like the stationary-object mass in our sampled scenes. This is because our generators sample initial conditions so that every body participates in the dynamics, and Sec. 4.1 already discussed removing non-physical entries (scene metadata, geometry, color, and g) from scoring. We note one more honest limitation: pixel-space MSE is a proxy for trajectory deviation and conflates visual salience with dynamical consequence, so the stratification should be read as ordinal rather than as a calibrated sensitivity coefficient.

\subsection{Additional Related Work Details}

\label{app:additional_related_work}

This appendix provides additional context for the related work summarized in Section~\ref{lit_review}. The main paper focuses on the closest conceptual gaps, while this section gives more detailed descriptions of prior benchmarks, evaluation protocols, and simulator-based approaches.

\subsubsection{Physical Property Estimation and Law Discovery from Video}

\label{app:law_discovery}

Physics 101~\citep{wu2016physics101} builds on the earlier Galileo framework~\citep{wu2015galileo}, which couples a physics engine with deep networks to infer object properties from video. Physics 101 is an early benchmark for learning physical properties from unlabeled video, compiling 17{,}408 clips across 101 objects that vary in material, shape, color, and size, and undergo four types of physical interaction: sliding, collision, spring attachment, and free fall. Related work has explored block-tower stability prediction~\citep{lerer2016tower}, graph-based dynamics learners that propagate object interactions~\citep{battaglia2016interaction,watters2017vin}, and inverse-graphics formulations that recover physical parameters by re-rendering~\citep{jaques2020pig}.

Visual Physics~\citep{chari2019visualphysics} takes a more direct approach to physics extraction from video: given bounding-box trajectories automatically extracted from video sequences, the system jointly discovers both the functional form of governing equations and their associated parameter values, without prior knowledge of which physical laws are operative. A complementary line of recent work jointly recovers 3D geometry and physical parameters from video by combining radiance fields or Gaussian splats with differentiable continuum mechanics~\citep{li2023pacnerf,xie2024physgaussian,zhong2024springgaus,cao2024neuma,cai2024gic}. These methods are closely related in spirit to our goal of recovering physics from visual observations, but the recovered representations are typically equations, parameters, trajectories, or implicit per-particle dynamics models rather than self-contained, re-runnable simulation programs.

\subsubsection{Benchmarks for Physical Understanding}

\label{app:physics_benchmarks}

PhysInOne~\citep{zhou2026physinone} is a large-scale benchmark for visual physics understanding, constructing 153{,}810 dynamic 3D scenes and 2 million videos across 71 physical phenomena, including rigid-body mechanics, fluid dynamics, optics, and electromagnetism. It uses multiple physics engines: Unreal Engine 5's Chaos Physics for rigid bodies, the Material Point Method~\citep{jiang2016mpm,hu2018mlsmpm} implemented with Taichi~\citep{hu2019taichi} for granular and deformable objects, and Smoothed Particle Hydrodynamics~\citep{monaghan1992sph} for liquids. Despite its scale, PhysInOne is a perception and estimation benchmark: models are evaluated on their ability to classify phenomena, estimate physical properties such as mass or elasticity, or predict future states.

Morpheus~\citep{zhang2025morpheus} evaluates whether contemporary video generative models, including SORA~\citep{brooks2024sora}, COSMOS~\citep{nvidia2025cosmos}, and WAN-2.1~\citep{wan2025wan}, preserve physical invariants. Rather than relying only on trajectory matching or subjective human ratings, Morpheus grounds evaluation in conservation laws: a Dynamical Score uses physics-informed neural networks~\citep{raissi2019pinn} to assess trajectory consistency with governing differential equations, while a Physical Invariance Score measures conservation of total energy and momentum across generated videos. VideoPhy~\citep{bansal2024videophy} instead focuses on physical commonsense in text-to-video generation, emphasizing interactions between materials of different types and using both human annotation and an automatic VideoCon-Physics evaluator, a VLM fine-tuned to assess physical plausibility from text-video pairs.

Beyond these closely related efforts, a broader benchmarking ecosystem evaluates physical understanding from complementary angles. VBench~\citep{huang2024vbench} evaluates video-generation quality; PhysBench~\citep{chow2025physbench}, PhyBench~\citep{qiu2024phybench}, and ComPhy~\citep{chen2022comphy} evaluate physical-commonsense reasoning in VLMs or LLMs; and Physion~\citep{bear2021physion}, IntPhys~\citep{riochet2018intphys}, CLEVRER~\citep{yi2020clevrer}, and ComPhy~\citep{chen2022comphy} study visual intuitive physics. These benchmarks are complementary to ours, but they generally evaluate physical understanding through classification, multiple-choice reasoning, plausibility judgment, or future-state prediction rather than executable program recovery.

\subsubsection{Physics Reasoning via Simulation}

\label{app:simulation_reasoning}

Solving Physics Olympiad via RL on Physics Simulators~\citep{prabhudesai2026physolym} addresses the scarcity of physics-reasoning training data relative to mathematics. The central idea is that physics simulators such as MuJoCo~\citep{todorov2012mujoco} can serve as procedurally scalable sources of supervision by randomizing physically valid scene graphs along a domain-specific language and querying numerical, inverse, and symbolic relationships between physical quantities. This line of work demonstrates the value of simulator-generated supervision for physics reasoning.

A parallel line of work develops differentiable simulators that propagate gradients through dynamics, including DiffTaichi~\citep{hu2020difftaichi}, Brax~\citep{freeman2021brax}, and Genesis~\citep{genesis2024}, with related infrastructure such as Isaac Gym~\citep{makoviychuk2021isaac} and NeuralSim~\citep{heiden2021neuralsim} used for robotics learning and system identification. Our work is related in that we also use physics simulators to generate ground-truth data, but differs in focusing on the video-conditioned inverse problem: given only a visual observation of a physical process, a model must recover explicit laws, parameters, and executable simulation code.

\subsection{CodeBLEU Specification}
Computed via the codebleu==0.7.0 PyPI package, weights=(0.25, 0.25, 0.25, 0.25), equal weights across the four components (n-gram match, weighted n-gram match with keyword upweighting, AST syntax match, dataflow match), default 4-gram BLEU with default Python keyword list, and the package's default Python tokenizer (Python's tokenize module).

\subsection{LLM Inter-rater Reliability}
Analyzing the stored justification fields of LLM judgment, we find 86\% of judgments agree within 1 unit and only 1.9\% show a spread $\geq 3$; the substantive disagreements cluster on coupled multi-body problems (e.g. \texttt{wedge\_block}, \texttt{wedge\_sphere}, \texttt{spring\_atwood}, \texttt{double\_pendulum}, \texttt{differential\_pulley\_chain\_fall}) where the ground truth uses one valid formulation (typically Lagrangian) but the candidate reaches for another (typically Newtonian). GPT-5-Nano tends to treat these as ``same physics, different formalism" (higher scores), while Gemini-2.5-Flash treats them as distinct theoretical frameworks (lower scores). Judge disagreement is also higher when scoring models whose outputs are vague-but-partially-correct (Gemma-3-12B mean spread $\approx$ 1.0) than models whose outputs are unambiguously wrong (Pixtral-12B mean spread $\approx$ 0.6).

% \subsection{Open-Source Model Coverage and Failures}
% \label{sec:oss-failures}

\subsection{Models Excluded From Evaluation}
\begin{itemize}
\itemsep0pt
\item \textbf{LLaVA-OneVision-1.5-8B-Instruct} (\texttt{lmms-lab}). vLLM~0.11.0 has no registered architecture for it; the bundled HF code requires a \texttt{transformers~5.x} build that is incompatible with vLLM in our environment. Cannot be loaded.
\item \textbf{GLM-4.5V} (\texttt{zai-org}). Listed nominally as a 9B-class model; the actual repository is a $>100$B-parameter MoE that exceeds both our disk budget and a single H200's memory. Excluded as out of scope.
\item \textbf{DeepSeek-VL2-Small} (\texttt{deepseek-ai}). Pipeline runs, but the model parrots prompt placeholders verbatim and never closes the \texttt{@@COT\_END@@} block before exhausting its output budget. Throughput is also $\sim$213\,s/sample ($\sim$144 hours per benchmark on one H200).
\item \textbf{VideoLLaMA3-7B} (\texttt{DAMO-NLP-SG}). 100\% refusal rate. Every output is the string \emph{``I'm sorry, but I can't assist with that.''} The safety post-training rejects long structured reasoning-plus-code prompts. The refusal itself is a reportable finding.
\item \textbf{LLaVA-OneVision-7B} (\texttt{lmms-lab}, \texttt{llava-hf}). Pipeline runs but every output degenerates into a 4096-token loop of \texttt{import matplotlib.pyplot as plt} repeated; the structured CoT block is never produced.
\end{itemize}

\subsection{Issues Affecting Included Models (Mitigated)}
\begin{itemize}
\itemsep0pt
\item \textbf{Per-model frame budgets are unequal.} Every video is decoded at 10 FPS, duration-preserving, and resized to 360×180 (SciPy) or 240×240 (Kubric). Each model then receives the largest number of frames its context window allows, sampled uniformly across the full clip. Gemini 2.5 Pro consumes the raw .mp4 at the same FPS and resolution. The effective sampling rate depends on clip duration, for a 12 s SciPy clip, budgets of 8/16/32/60/80 frames correspond to about 0.7/1.3/2.7/5.0/6.7 FPS; for a 5 s Kubric clip the 10 FPS decode caps the pool at 50 frames. We treat the closed-vs-open comparison as confounded by frame budget; within-OSS comparisons are cleanly shown in Table~\ref{tab:oss-results}.
\item \textbf{GLM-4V-9B is single-image-only in vLLM and capped at 8K positional embeddings.} We tile the sampled frames into a single 4-column composite image; we also capped \texttt{max\_tokens} at 4096 (with $\sim$2K-token prompts), which contributes truncation-not-malformation to its CoT-parse failures, particularly on the deeper kubric schema.
\end{itemize}

\begin{table}[t]
\centering
\footnotesize
\setlength{\tabcolsep}{0.4em}
\caption{\textbf{Open-source model parse rates on the 2,430-sample benchmark.} \textbf{CoT} = fraction of outputs whose \texttt{@@COT\_BEGIN@@} block contains parseable JSON; \textbf{Code} = fraction whose \texttt{@@CODE\_BEGIN@@} block contains a non-empty Python program. Code metric is a structural check, not an executability check. $^\dagger$\,Tiled into a single 4-column composite image (vLLM single-image limit).}
\label{tab:oss-results}
\begin{adjustbox}{width=0.75\linewidth}
\begin{tabular}{l|cc|cc|c}
\toprule
\textbf{Model} & \multicolumn{2}{c|}{\textbf{scipy}} & \multicolumn{2}{c|}{\textbf{kubric}} & \textbf{Frames} \\
& \textbf{CoT} & \textbf{Code} & \textbf{CoT} & \textbf{Code} & per video \\
\midrule
GLM-4V-9B                 & 67.1\%          & 94.0\%          & 47.1\%          & 70.5\%          & 16$^\dagger$ \\
Pixtral-12B-2409          & 99.6\%          & 99.8\%          & 82.1\%          & 87.1\%          & 24 \\
InternVL3.5-30B-A3B (MoE) & 99.2\%          & 99.8\%          & \textbf{100.0\%} & 93.2\%          & 32 \\
Qwen3-VL-30B-A3B (MoE)    & \textbf{99.9\%} & 94.2\%          & 99.9\%          & 85.3\%          & 60 \\
GLM-4.1V-9B-Thinking      & 88.2\%          & 96.6\%          & 95.6\%          & 85.5\%          & 60 \\
Gemma-3-12B-IT & 60.5\% & \textbf{99.9\%} & 79.7\% & \textbf{100.0\%} & 32 \\
\bottomrule
\end{tabular}
\end{adjustbox}
\end{table}

\subsection{More Dataset Statistics}
\label{app_stat}
For Scipy experiments that deal with ordinary differential equations (ODEs), the Dormand–Prince RK45 scheme of SciPy is used with tight tolerances (\texttt{rtol} = $10^{-9}$, \texttt{atol} = $10^{-11}$) consistently across all simulation types. The resulting trajectories are rendered into animations using Matplotlib’s \texttt{FuncAnimation} and encoded as MP4 videos at 30 FPS over a 12-second duration using FFmpeg. The physics of the Kubric experiments is simulated by PyBullet in headless mode at 240\,Hz with 8 substeps per render frame, and rendering uses \texttt{pyrender} and \texttt{trimesh} under EGL-accelerated OpenGL with PBR materials. The resulting trajectories are rendered as MP4 videos.

The CoT field \texttt{ode\_system.equations\_latex} contains a median of 3
\LaTeX{} equations per sample (range 1--22) and \texttt{derivation\_steps} contains a median of 4 ordered steps from first principles to the implemented update (range 3--47). The total annotation corpus comprises \textbf{1{,}112{,}171 derivation steps} and \textbf{588{,}023 \LaTeX{} equations} across 160,614 video--code--CoT samples, providing dense supervision for both equation-recovery and step-by-step reasoning evaluations. The annotation set covers \textbf{154 distinct named physical laws} (86 in Pipeline~A, 68 in Pipeline~B), spanning Newton's second law, Lagrangian mechanics, rigid-body Newton-Euler dynamics, Coulomb friction, restitution-impulse contact, wave equations, constrained dynamics, and rotational kinematics, among others. For experiments whose internal randomization is governed by a single integer seed leading to many choices (categorical object types, variable object counts $N \in [2, 10]$), we manually elevate the sample count proportionally to the effective hidden dimensionality.

Across all 160{,}614 samples we ship \textbf{21.4 million lines of
executable Python} (mean 134 LOC, median 121 LOC per sample), each of which is a self-contained, runnable simulation script. The
\texttt{parameters} field contains a structured numerical record with a median of 9 and up to 21 typed parameters per sample, totaling \textbf{1.54M parameter values} suitable for parameter-estimation benchmarks. On disk, the dataset occupies 53.9 GB (49.8 GB SciPy video + 1.96 GB Kubric video + 1.08 GB code + 0.88 GB CoT JSON + 0.15 GB parameter JSON).

\begin{figure}[ht]
\centering
\includegraphics[width=\linewidth]{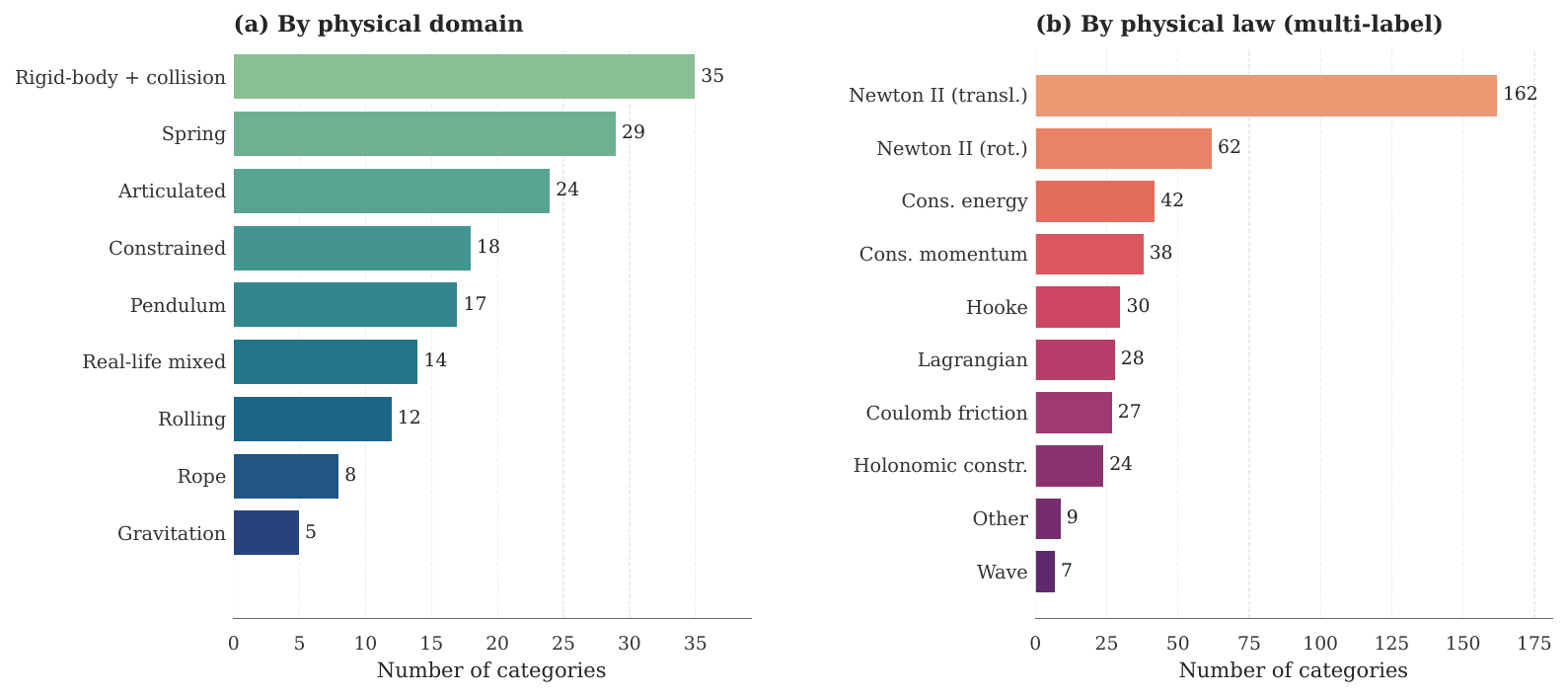}
\caption{Distribution of the 162 categories.
  \textbf{(a)} Nine top-level physical domains.
  \textbf{(b)} Physical laws invoked across the dataset (multi-label, so
  counts sum to more than 162); Newton's 2nd law appears in every
  category by construction, while the long tail captures laws like the
  1-D damped wave equation, gravitation, hydrostatic equilibrium, and
  the Lorentz force.}
\label{fig:dist}
\end{figure}

The 154 distinct physical-law annotations span the standard
undergraduate to early-graduate physics curriculum: classical mechanics (Newton's laws, Lagrangian/Hamiltonian formulations), rigid-body dynamics with contact (Newton-Euler equations, Coulomb friction, restitution-impulse models), oscillatory and wave systems (linear and nonlinear pendula, coupled oscillators, 1-D wave equations with damping), constrained dynamics (Atwood machines, pulleys, ropes), and fluid-coupled systems (granular flow, sloshing, sphere-basin oscillations). Figure \ref{fig:dist} shows the categorical counts as per physical laws and physical domain. Each law name is paired with a formal statement and
\LaTeX{} formula, providing a labeled, grounded ontology for evaluating
whether a model identifies the \emph{correct} physics regime in addition to extracting the correct numerical parameters.

\subsection{Object Count per Experiment Category}
Object count per experiment is an indicator of the complexity of that experiment. We computed this from the released params.json of all 162 categories heuristically (because sometimes defining an object is very difficult, read below). A script counted independently simulated dynamical entities (rigid bodies and point masses; ropes/chains by their links), excluding static scenery and the ground plane.

\begin{table}[t]
\centering
\begin{tabular}{lrrrr}
\toprule
\textbf{Objects per category} & $\leq 2$ & $2$--$5$ & $5$--$10$ & $>10$ \\
\midrule
SciPy  & 61 & 9  & 1  & 0 \\
Kubric & 38 & 14 & 15 & 7 \\
\bottomrule
\end{tabular}
\end{table}

Per sample we read an explicit count field (like number of balls, pit, dominoes, etc) where one exists, else the length of the per-object arrays (objects, masses, object shapes), else the number of distinct mass parameters; we then average over a category's samples. This table counts categories, not samples.

For 17 categories an object count is genuinely undefined, which is itself informative: in \texttt{brachistochrone\_race} and \texttt{hemisphere\_roll\_off} the dynamics are mass-independent, so no mass is ever sampled; \texttt{accelerating\_beaker\_water}, \texttt{u\_tube\_fluid\_oscillation} and \texttt{wave\_interference} are continua (fluid columns, strings with linear density $\mu$) rather than discrete bodies; and in \texttt{three\_gear\_chain}, $N_A = 30$ counts gear teeth, not objects. We would caution, though, that object count is a weak proxy for the complexity a rater must track. A double pendulum has only two bobs but is chaotic and requires coupled nonlinear equations, which is complex, whereas two blocks colliding on a plane involves the same object count and a single impulse, which is much simpler. Object count also cannot be equated with the number of masses, because several of our categories depend on massless elements that are nonetheless dynamically essential, springs, strings, tethers and pulley constraints that transmit forces without contributing inertia. This is why the complexity statistic we already report is \texttt{n\_rand}, the number of independently randomized physical parameters per category (Tables \ref{tab:kubric_per_experiment} and \ref{tab:scipy_per_experiment}), which drives our per-category sample budget; and for categories where a single integer seed determines many latent choices at once (object type, object count, material), \texttt{n\_rand} understates the true dimensionality, which is why we manually elevated their sample counts (Sec. 3.2).

\begin{longtable}{lrrc}
\caption{Pipeline~A (SciPy): per-experiment statistics. $\star$ marks seed-based experiments where a single integer seed drives non-factorizable hidden choices, requiring manually-elevated sample counts.}
\label{tab:scipy_per_experiment}\\
\toprule
\textbf{Experiment} & \textbf{$n_{\text{rand}}$} & \textbf{Samples} & \textbf{Seeded} \\
\midrule
\endfirsthead
\toprule
\textbf{Experiment} & \textbf{$n_{\text{rand}}$} & \textbf{Samples} & \textbf{Seeded} \\
\midrule
\endhead
\bottomrule
\multicolumn{4}{r}{\textit{(continued on next page)}} \\
\endfoot
\bottomrule
\textbf{Total} (87 experiments) & & \textbf{64,759} & \\
\bottomrule
\endlastfoot
  \texttt{complex\_collision\_plane} & 2 & 2500 & $\star$ \\
  \texttt{accelerating\_cart\_sliding\_mass} & 4 & 2000 & $\star$ \\
  \texttt{balls\_in\_basin} & 4 & 2000 & $\star$ \\
  \texttt{disk\_spinoff} & 7 & 1500 & $\star$ \\
  \texttt{block\_on\_sphere\_basin} & 13 & 1300 &  \\
  \texttt{molecule\_2} & 12 & 1200 &  \\
  \texttt{three\_body\_gravity} & 11 & 1100 &  \\
  \texttt{hammer\_projectile\_bounce} & 10 & 1000 &  \\
  \texttt{rolling\_race\_hoop\_vs\_disk\_vs\_sphere} & 4 & 1000 & $\star$ \\
  \texttt{springs\_two\_slabs\_compound} & 10 & 1000 &  \\
  \texttt{toppling\_cylinder\_wedge} & 10 & 959 &  \\
  \texttt{two\_conveyor\_drop} & 10 & 1000 &  \\
  \texttt{wedge\_block} & 10 & 1000 &  \\
  \texttt{wheel\_down\_stairs} & 10 & 1000 &  \\
  \texttt{cart\_two\_metronomes} & 9 & 900 &  \\
  \texttt{double\_2d\_spring} & 9 & 900 &  \\
  \texttt{falling\_spring} & 9 & 900 &  \\
  \texttt{moveable\_double\_pendulum} & 9 & 900 &  \\
  \texttt{rod\_ball\_strike} & 9 & 900 &  \\
  \texttt{spinning\_disc\_block\_spring} & 9 & 900 &  \\
  \texttt{wheel\_belt\_drop} & 9 & 900 &  \\
  \texttt{cart\_pendulum} & 8 & 800 &  \\
  \texttt{dangle\_stick} & 8 & 800 &  \\
  \texttt{differential\_pulley\_chain\_fall} & 8 & 800 &  \\
  \texttt{discs\_friction\_coupling} & 8 & 800 &  \\
  \texttt{pendulum\_block\_strike} & 8 & 800 &  \\
  \texttt{slab\_springs\_block\_drop} & 8 & 800 &  \\
  \texttt{sliding\_block\_step\_topple} & 8 & 800 &  \\
  \texttt{spinning\_ball\_bounce} & 8 & 800 &  \\
  \texttt{spring\_2d} & 8 & 800 &  \\
  \texttt{tilting\_wedge\_cylinder} & 8 & 800 &  \\
  \texttt{wheel\_wall\_bounce} & 8 & 800 &  \\
  \texttt{block\_strikes\_spring\_pair} & 7 & 700 &  \\
  \texttt{bouncing\_rod} & 7 & 700 &  \\
  \texttt{chain\_of\_springs} & 7 & 700 &  \\
  \texttt{cylinder\_piston\_ball} & 7 & 700 &  \\
  \texttt{double\_spring} & 7 & 700 &  \\
  \texttt{gear\_rack\_spring\_oscillator} & 7 & 700 &  \\
  \texttt{half\_cylinder\_drop} & 7 & 700 &  \\
  \texttt{magnet\_wheel} & 7 & 700 &  \\
  \texttt{ring\_sticky\_collision} & 7 & 700 &  \\
  \texttt{spinning\_ball\_on\_moving\_belt} & 7 & 700 &  \\
  \texttt{spring\_mass\_string\_fixed} & 7 & 700 &  \\
  \texttt{spring\_mass\_string\_ring} & 7 & 700 &  \\
  \texttt{stacked\_balls\_drop} & 7 & 700 &  \\
  \texttt{string\_ring\_wave} & 7 & 700 &  \\
  \texttt{three\_gear\_chain} & 7 & 700 &  \\
  \texttt{three\_gear\_middle\_drive} & 7 & 700 &  \\
  \texttt{wedge\_sphere} & 7 & 700 &  \\
  \texttt{ballistic\_pendulum} & 6 & 600 &  \\
  \texttt{belt\_gear\_rack\_train} & 3 & 600 & $\star$ \\
  \texttt{billiards\_break} & 6 & 600 &  \\
  \texttt{block\_chain\_collision} & 6 & 600 &  \\
  \texttt{colliding\_blocks} & 6 & 600 &  \\
  \texttt{double\_pendulum} & 6 & 600 &  \\
  \texttt{driven\_string\_fixed} & 6 & 600 &  \\
  \texttt{driven\_string\_ring} & 6 & 600 &  \\
  \texttt{hemisphere\_roll\_off} & 6 & 600 &  \\
  \texttt{moveable\_pendulum} & 6 & 600 &  \\
  \texttt{projectile\_bounce} & 6 & 600 &  \\
  \texttt{rigid\_double\_pendulum} & 6 & 600 &  \\
  \texttt{roller\_single} & 6 & 600 &  \\
  \texttt{spool\_pull\_direction\_2d} & 6 & 600 &  \\
  \texttt{string\_block\_wave} & 6 & 600 &  \\
  \texttt{two\_gear\_belt} & 6 & 600 &  \\
  \texttt{u\_tube\_fluid\_oscillation} & 6 & 600 &  \\
  \texttt{watt\_straight\_line\_linkage} & 6 & 600 &  \\
  \texttt{wave\_interference} & 6 & 600 &  \\
  \texttt{accelerating\_beaker\_water} & 5 & 500 &  \\
  \texttt{block\_belt\_spring\_stickslip} & 5 & 500 &  \\
  \texttt{conveyor\_belt\_drop} & 5 & 500 &  \\
  \texttt{robot\_speed} & 5 & 500 &  \\
  \texttt{semicircular\_track} & 5 & 500 &  \\
  \texttt{simple\_pendulum} & 5 & 500 &  \\
  \texttt{single\_spring} & 5 & 500 &  \\
  \texttt{two\_carts\_slack\_string} & 5 & 500 &  \\
  \texttt{atwood\_on\_table} & 4 & 400 &  \\
  \texttt{ball\_wrapping\_pole} & 4 & 400 &  \\
  \texttt{brachistochrone} & 4 & 400 &  \\
  \texttt{lagrange\_roller} & 4 & 400 &  \\
  \texttt{ring\_pendulum} & 4 & 400 &  \\
  \texttt{spring\_atwood} & 4 & 400 &  \\
  \texttt{two\_body\_gravity} & 4 & 400 &  \\
  \texttt{compound\_atwood} & 3 & 300 &  \\
  \texttt{sand\_onto\_moving\_cart} & 3 & 300 &  \\
  \texttt{brachistochrone\_race} & 2 & 200 &  \\
  \texttt{pendulum\_slack\_revolution} & 2 & 200 &  \\
\end{longtable}

\begin{longtable}{lrccr}
\caption{Pipeline~B (Kubric): per-experiment statistics. $\bullet$ in \emph{Var} marks experiments with variable object counts (boost $\times 1.5$); $\bullet$ in \emph{Col} marks experiments with random color/material choice (boost $\times 1.3$). Sample counts are floored at 400 and capped at 5{,}000.}
\label{tab:kubric_per_experiment}\\
\toprule
\textbf{Experiment} & \textbf{$n_{\text{rand}}$} & \textbf{Var} & \textbf{Col} & \textbf{Samples} \\
\midrule
\endfirsthead
\toprule
\textbf{Experiment} & \textbf{$n_{\text{rand}}$} & \textbf{Var} & \textbf{Col} & \textbf{Samples} \\
\midrule
\endhead
\bottomrule
\multicolumn{5}{r}{\textit{(continued on next page)}} \\
\endfoot
\bottomrule
\textbf{Total} (75 experiments) & & & & \textbf{95,855} \\
\bottomrule
\endlastfoot
  \texttt{multi\_physics\_arena} & 47 & $\bullet$ &  & 5000 \\
  \texttt{tumbling\_dice} & 19 & $\bullet$ & $\bullet$ & 3705 \\
  \texttt{bouncing\_balls\_arena} & 16 & $\bullet$ & $\bullet$ & 3120 \\
  \texttt{balloon\_tether\_wall\_pull} & 15 & $\bullet$ & $\bullet$ & 2925 \\
  \texttt{turntable\_mirror\_blocks} & 12 & $\bullet$ & $\bullet$ & 2340 \\
  \texttt{rope\_pendulum\_swing} & 14 & $\bullet$ &  & 2100 \\
  \texttt{ball\_pit\_drop} & 10 & $\bullet$ & $\bullet$ & 1950 \\
  \texttt{block\_tower\_projectile} & 13 & $\bullet$ &  & 1950 \\
  \texttt{block\_wall\_hit\_by\_tethered\_mass} & 10 & $\bullet$ & $\bullet$ & 1950 \\
  \texttt{rod\_spin\_hits\_balloon} & 14 &  & $\bullet$ & 1820 \\
  \texttt{jenga\_pull} & 12 & $\bullet$ &  & 1800 \\
  \texttt{accelerating\_panel\_spins\_objects} & 9 & $\bullet$ & $\bullet$ & 1755 \\
  \texttt{avalanche\_slope} & 9 & $\bullet$ & $\bullet$ & 1755 \\
  \texttt{granular\_hourglass\_3d} & 9 & $\bullet$ & $\bullet$ & 1755 \\
  \texttt{concave\_bowl\_spin\_orbit} & 12 &  & $\bullet$ & 1560 \\
  \texttt{fan\_push\_stationary\_block} & 8 & $\bullet$ & $\bullet$ & 1560 \\
  \texttt{moving\_cart\_hits\_moving\_cart} & 12 &  & $\bullet$ & 1560 \\
  \texttt{spring\_compress\_vs\_stretch} & 12 &  & $\bullet$ & 1560 \\
  \texttt{scissor\_collapse} & 10 & $\bullet$ &  & 1500 \\
  \texttt{stacked\_discs\_topple} & 10 & $\bullet$ &  & 1500 \\
  \texttt{staircase\_bounce} & 10 & $\bullet$ &  & 1500 \\
  \texttt{carried\_pendulum\_inertia} & 11 &  & $\bullet$ & 1430 \\
  \texttt{crank\_push\_friction\_stop} & 11 &  & $\bullet$ & 1430 \\
  \texttt{lever\_catapult} & 14 &  &  & 1400 \\
  \texttt{trebuchet\_throw} & 14 &  &  & 1400 \\
  \texttt{domino\_chain} & 9 & $\bullet$ &  & 1350 \\
  \texttt{funnel\_sorting} & 9 & $\bullet$ &  & 1350 \\
  \texttt{wrecking\_ball\_pendulum} & 9 & $\bullet$ &  & 1350 \\
  \texttt{moving\_cart\_hits\_stationary\_cart} & 10 &  & $\bullet$ & 1300 \\
  \texttt{ramp\_to\_spring\_projectile} & 10 &  & $\bullet$ & 1300 \\
  \texttt{marble\_run\_ramps} & 8 & $\bullet$ &  & 1200 \\
  \texttt{robot\_arm\_collapse} & 8 & $\bullet$ &  & 1200 \\
  \texttt{mass\_spring\_collision\_chain} & 6 & $\bullet$ & $\bullet$ & 1170 \\
  \texttt{spinning\_top\_precession} & 9 &  & $\bullet$ & 1170 \\
  \texttt{stacked\_blocks\_earthquake} & 6 & $\bullet$ & $\bullet$ & 1170 \\
  \texttt{toppling\_bookshelf\_row} & 6 & $\bullet$ & $\bullet$ & 1170 \\
  \texttt{wind\_balanced\_falling\_panel} & 6 & $\bullet$ & $\bullet$ & 1170 \\
  \texttt{gyroscope\_on\_stand} & 8 &  & $\bullet$ & 1040 \\
  \texttt{spinning\_rod\_balloon\_tether} & 8 &  & $\bullet$ & 1040 \\
  \texttt{two\_car\_collision\_3d} & 8 &  & $\bullet$ & 1040 \\
  \texttt{object\_collision\_mid\_air} & 10 &  &  & 1000 \\
  \texttt{arch\_collapse\_keystone} & 5 & $\bullet$ & $\bullet$ & 975 \\
  \texttt{bridge\_truss\_load\_failure} & 5 & $\bullet$ & $\bullet$ & 975 \\
  \texttt{coin\_toss\_spin\_land} & 7 &  & $\bullet$ & 910 \\
  \texttt{seesaw\_with\_falling\_mass} & 7 &  & $\bullet$ & 910 \\
  \texttt{springboard\_rebound\_ball} & 7 &  & $\bullet$ & 910 \\
  \texttt{vertical\_fall\_spinning\_panel} & 7 &  & $\bullet$ & 910 \\
  \texttt{newton\_cradle\_3d} & 6 & $\bullet$ &  & 900 \\
  \texttt{plasticine\_ball\_inelastic\_drop} & 9 &  &  & 900 \\
  \texttt{projectile\_parabola} & 9 &  &  & 900 \\
  \texttt{ragdoll\_free\_fall} & 9 &  &  & 900 \\
  \texttt{ramp\_slide\_friction} & 9 &  &  & 900 \\
  \texttt{rolling\_cylinder\_slope} & 9 &  &  & 900 \\
  \texttt{seesaw\_imbalance} & 9 &  &  & 900 \\
  \texttt{wedge\_collision} & 9 &  &  & 900 \\
  \texttt{bowling\_strike} & 8 &  &  & 800 \\
  \texttt{double\_pendulum\_3d\_chaotic} & 8 &  &  & 800 \\
  \texttt{ramp\_jump\_landing} & 8 &  &  & 800 \\
  \texttt{sphere\_bowl\_oscillation} & 8 &  &  & 800 \\
  \texttt{conical\_pendulum} & 6 &  & $\bullet$ & 780 \\
  \texttt{oblique\_projectile\_hits\_slope} & 6 &  & $\bullet$ & 780 \\
  \texttt{pendulum\_3d\_single} & 6 &  & $\bullet$ & 780 \\
  \texttt{rolling\_vs\_sliding\_blocks} & 6 &  & $\bullet$ & 780 \\
  \texttt{coin\_spin\_topple} & 7 &  &  & 700 \\
  \texttt{elastic\_coupled\_cart\_collision} & 7 &  &  & 700 \\
  \texttt{stick\_support\_catapult\_release} & 7 &  &  & 700 \\
  \texttt{compound\_pulley\_lift} & 5 &  & $\bullet$ & 650 \\
  \texttt{offset\_load\_turntable} & 5 &  & $\bullet$ & 650 \\
  \texttt{rolling\_ball\_loop\_track} & 5 &  & $\bullet$ & 650 \\
  \texttt{rolling\_ball\_up\_slope} & 5 &  & $\bullet$ & 650 \\
  \texttt{atwood\_machine} & 6 &  &  & 600 \\
  \texttt{double\_pendulum\_impact\_transfer} & 6 &  &  & 600 \\
  \texttt{hinged\_door\_slam} & 6 &  &  & 600 \\
  \texttt{center\_vs\_offset\_seesaw} & 5 &  &  & 500 \\
  \texttt{billiard\_break\_3d} & 4 &  &  & 400 \\
\end{longtable}

\subsection{Frame-count Ablation}
We wanted to see how changing the number of frames affects the model performance for the open-source models. So, we re-ran the full pipeline over 162 videos (1 from each experiment) × 5 frame budgets for Qwen3-VL-30B and 4 budgets for InternVL3.5-30B, and re-scored both the Claude and GPT arms with the parameter matcher. In Table~\ref{tab:frame_ablation}, * denotes the frame budget used in the paper. (N=162) per row, with 1,284 ground-truth parameters and zero inference errors.

We see that rankings are mostly frame-budget invariant. Qwen3-VL at its worst budget (5.76\%) still beats InternVL at its best (4.28\%). CodeBLEU is essentially constant. For Qwen, naming recall keeps rising with frames while value accuracy falls after 32. More frames enable the model to name more of the right parameters, but not better quantitative estimation. This supports the paper's central claim that the bottleneck is quantitative physical inference, not perception bandwidth.

\begin{table*}[t]
\centering
\footnotesize
\setlength{\tabcolsep}{2pt}
\caption{How the parameter estimation and coding performance of the two open-source models vary with different number of input frames for Qwen3-VL-30B and InternVL3.5-30B.}
\label{tab:frame_ablation}
\begin{tabular}{lcccccccc}
\toprule
Model & Frames & Param. Acc. & Naming Recall & Value Acc. & No Errors & Video Save & CodeBLEU & Prompt Tok. \\
\midrule
Qwen      & 8    & 5.84\%          & 0.384          & 0.155          & 78.4\%          & 82.7\%          & 0.203 & 2,327  \\
Qwen      & 16   & 6.23\%          & 0.387          & 0.165          & 75.3\%          & 80.9\%          & 0.206 & 2,863  \\
Qwen      & 32   & \textbf{6.31\%} & 0.399          & \textbf{0.168} & 71.0\%          & 72.8\%          & 0.206 & 3,937  \\
Qwen      & 60*  & 6.07\%          & \textbf{0.409} & 0.157          & 75.3\%          & 80.9\%          & 0.206 & 5,468  \\
Qwen      & 80   & 5.76\%          & 0.408          & 0.143          & 77.8\%          & 80.2\%          & 0.208 & 6,190  \\
\midrule
InternVL   & 8    & 3.12\%          & 0.247          & 0.126          & 48.1\%          & 65.4\%          & 0.194 & 3,905  \\
InternVL   & 16   & 3.43\%          & 0.256          & 0.132          & 49.4\%          & 64.2\%          & 0.196 & 6,016  \\
InternVL   & 32*  & 4.05\%          & 0.265          & 0.163          & \textbf{50.6\%} & \textbf{66.7\%} & 0.196 & 10,240 \\
InternVL   & 60   & \textbf{4.28\%} & \textbf{0.273} & \textbf{0.168} & 46.3\%          & 60.5\%          & 0.195 & 16,247 \\
\bottomrule
\end{tabular}
\end{table*}

\subsection{Additional Cohen-$\kappa$ Plots}
\begin{figure}[t]
    \centering

    \begin{subfigure}[t]{0.48\linewidth}
        \centering
        \includegraphics[width=\linewidth]{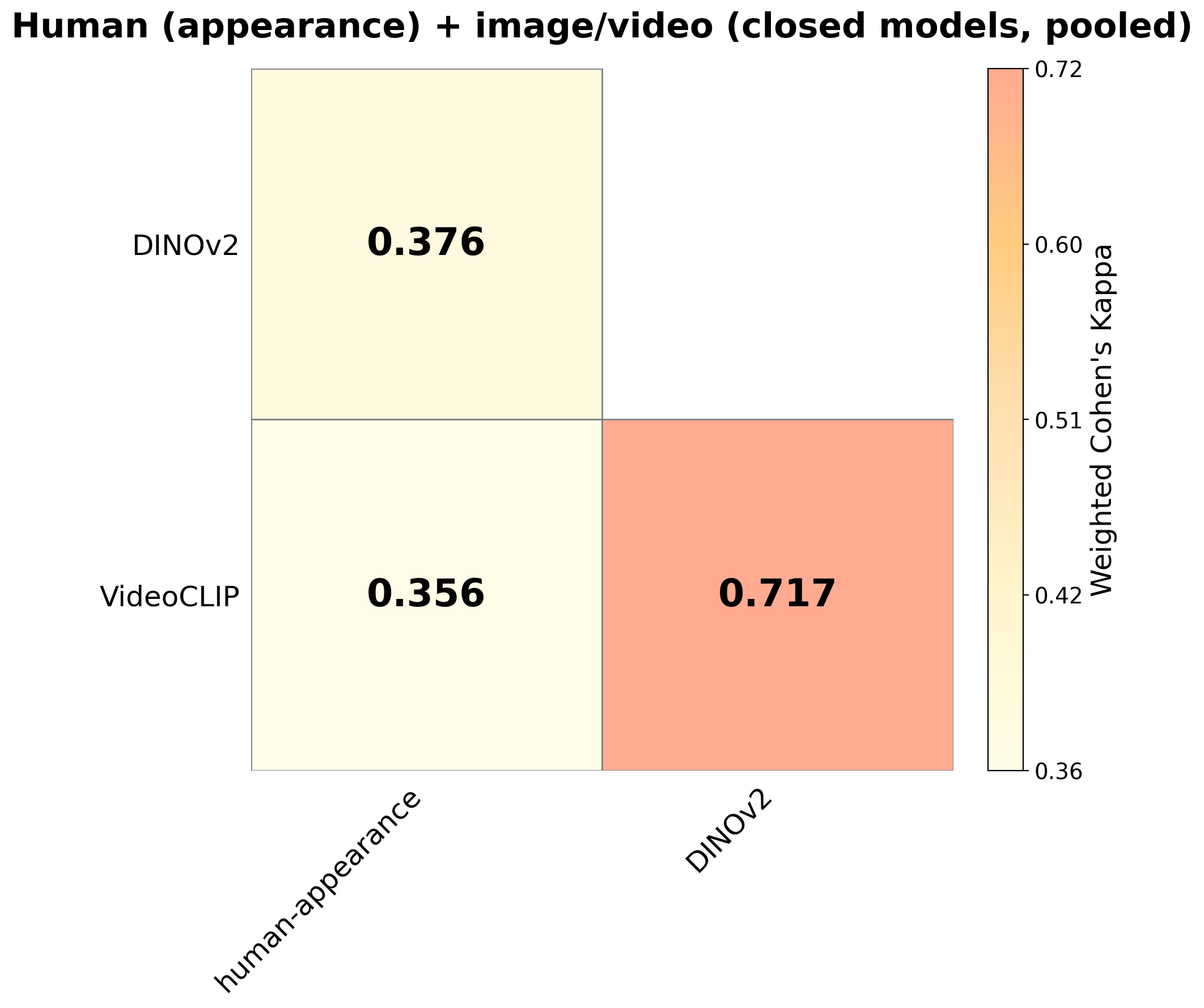}
        \caption{}
        \label{fig:a}
    \end{subfigure}
    \hfill
    \begin{subfigure}[t]{0.48\linewidth}
        \centering
        \includegraphics[width=\linewidth]{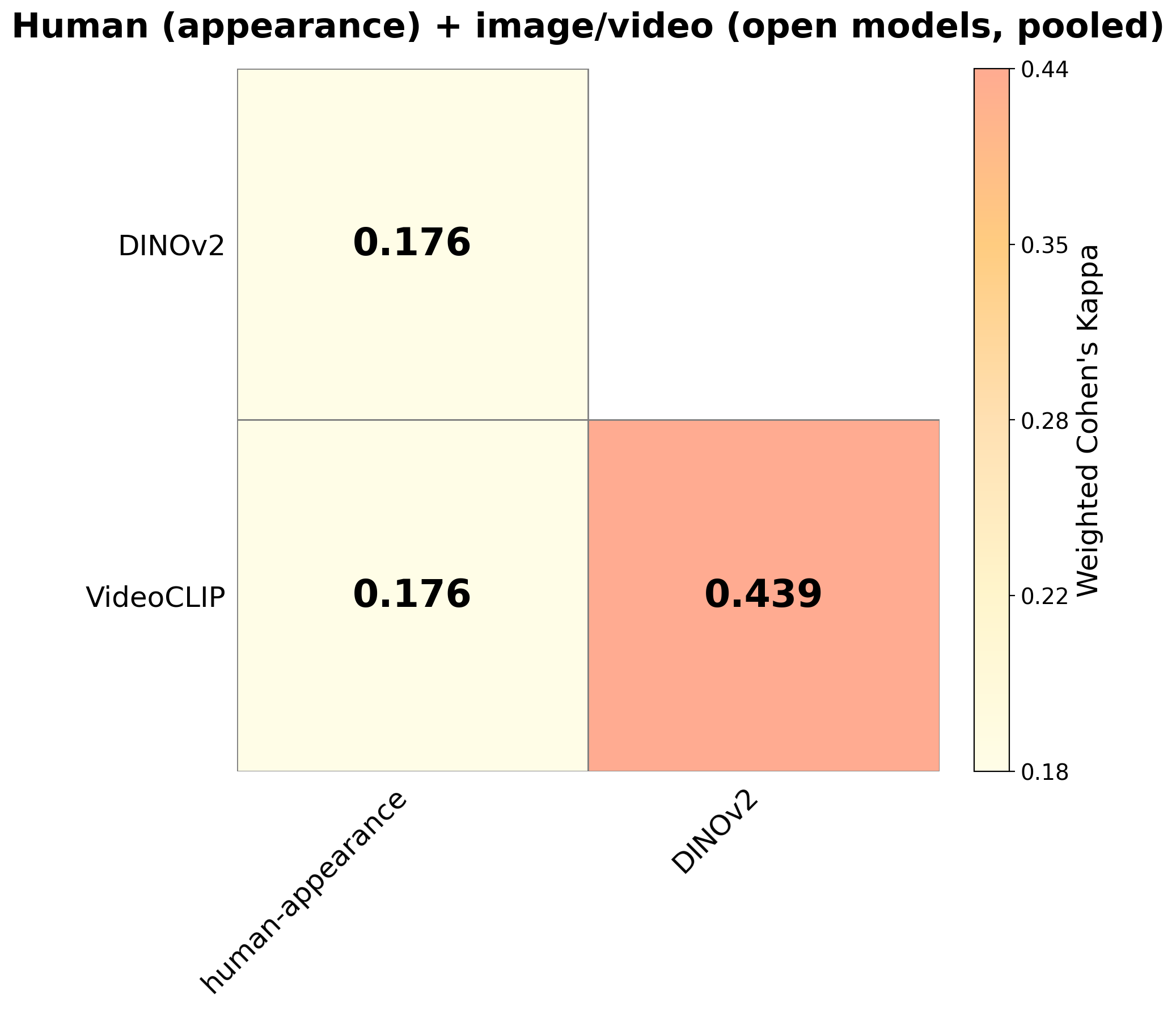}
        \caption{}
        \label{fig:b}
    \end{subfigure}

    \caption{Cohen's-$\kappa$ plot of human-rated video appearance similarity score with the verifiable patch-wise similarity metrics (DinoV2 and VideoCLIP) for (a) closed and (b) open-source models.}
    \label{fig:twoimages}
\end{figure}

Figure \ref{fig:twoimages} shows that the human-rated video appearance similarity score is less correlated with patchwise similarity, but the correlation is definitely much higher than Figure \ref{fig:four_images}'s human-rated physics similarity. The reason is clear. Both these patchwise similarity metrics and human (when asked to rate based on visual similarity) focus on the pixel level match rather than physics understanding.

\begin{figure}
    \centering
    \includegraphics[width=\linewidth]{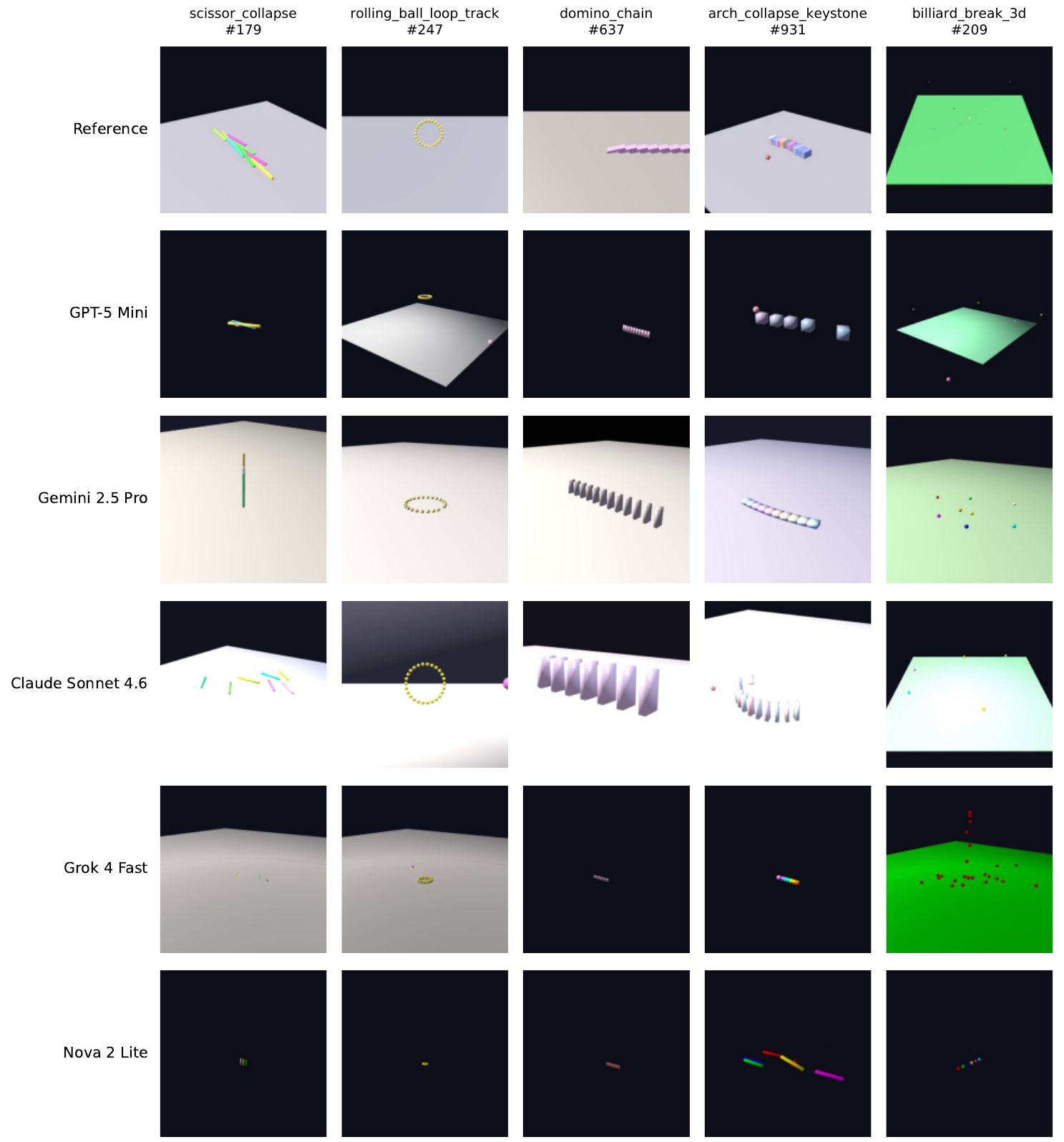}
    \caption{Videos rendered by code generated by closed-source models. The top row is the input (reference) video.}
    \label{fig:failure_closed}
\end{figure}

\begin{figure}
    \centering
    \includegraphics[width=\linewidth]{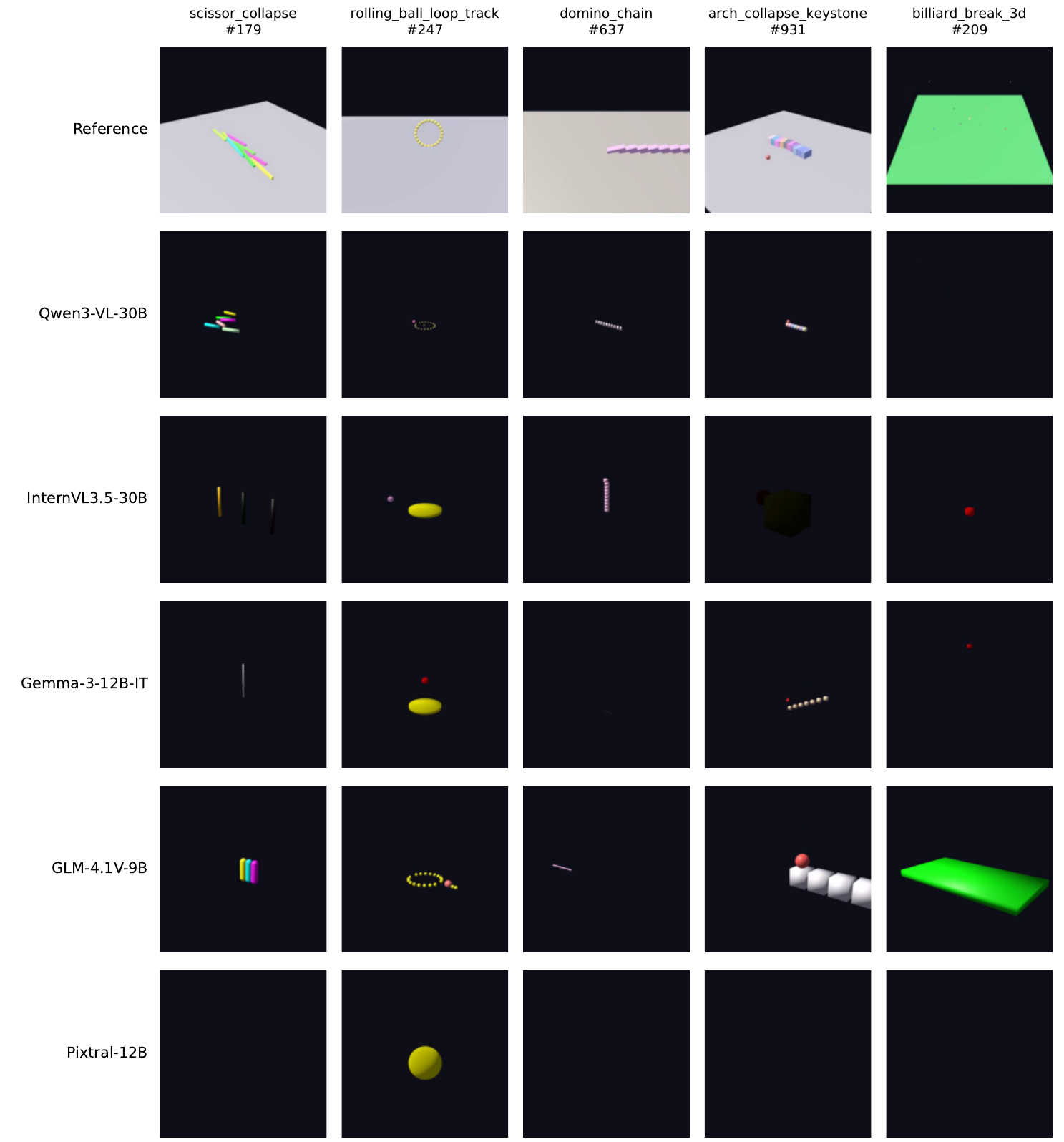}
    \caption{Videos rendered by code generated by open-source models. The top row is the input (reference) video.}
    \label{fig:failure_open}
\end{figure}

\subsection{Transfer to real world}
\label{app:transfer2real}
We scoped the paper to single-scene, short clips. But we think the synthetic setting is the right place to start. Our videos are the easy case: no sensor noise, no occlusion, clean lighting, and we even hand the model the engine name and an experiment-agnostic code template so it is not penalized for engine unfamiliarity. Models still fail badly: 2.47/5 (the best model) on inferring the demonstrated law, 7.7\% parameter recovery within ±20\%. A model that cannot recover friction from a clean render of a block on a ramp will not recover it from a noisy, partially occluded one. Reporting the failure in the absence of nuisance factors is more informative than reporting it in their presence, because it isolates the cause. Also, this measurement is possible synthetically. It’s hard to annotate real videos with ground-truth mass, stiffness or restitution, etc., so a ±20\% parameter-recovery metric cannot be defined on it easily.

We also observed two failures that are genuinely about physics rather than about guessing a simulator's generative parameters. (i) Models state physics laws more correctly than infer them from video. So the missing capability is binding perception to physics. (ii) Appearance is reproduced far better than dynamics; Figure 4 shows the patchwise metrics correlate strongly with each other but not with human physics judgment. Models match pixels and miss the dynamics, a physics gap.

\subsection{Benchmark Novelty}
The individual ingredients exist in literature, but no prior benchmark joins them. Prior work splits cleanly into two lines that have not met: property estimation from video (Zhan et al. 2025, ContPhy, Physion, ComPhy, PhysInOne) recovers a handful of latent values but never a program; physics code generation (PhysCodeBench, SimuScene) produces executable simulations but is text-conditioned and single-engine, so nothing is inferred from pixels. PhySimCode closes the loop: video to laws + parameters + runnable code to regenerated video scored back against the input. Table 1 shows we are the only row with Video, CoT, Code, Physics laws, Engine-agnostic, and 2D+3D all present, at 162 phenomena vs. 4–71 in prior benchmarks. We annotate the governing law and ground it in video. We can separate what a model knows from what it can infer: retrieval vs. grounding. We treat code as a lossless, executable representation of scientific visual media. So PhySimCode gives some novel and useful insights about MLLM’s physics understanding.

\end{document}